\documentclass{article}

\usepackage{arxiv}

\usepackage[utf8]{inputenc} 
\usepackage[T1]{fontenc}    
\usepackage{hyperref}       
\usepackage{url}            
\usepackage{booktabs}       
\usepackage{amsfonts}       
\usepackage{nicefrac}       
\usepackage{microtype}      
\usepackage{lipsum}
\usepackage{multicol}
\usepackage{algorithm2e}
\usepackage{subcaption}
\usepackage{calc}
\usepackage{amssymb}
\usepackage{amstext}
\usepackage{amsmath}
\usepackage{amsthm}
\usepackage{algorithmic}
\usepackage{textcomp}
\usepackage{xcolor}
\usepackage{multirow}
\usepackage{graphicx}
\graphicspath{ {./figures/} }

\title{DeepSeekMath Meets Order Book: Group-Aware Policy Optimization for High-Frequency Directional Trading}

\author{
 Sayak Chakrabarty$^\dagger$ \\
  Department of Computer Science\\
  Northwestern University\\
  Evanston, IL 60208, USA \\
  \texttt{sayakchakrabarty2025@u.northwestern.edu}
   \And
 Souradip Pal$^\dagger$ \\
  School of Electrical and Computer Engineering\\
  Purdue University\\
  West Lafayette, IN 47906, USA \\
  \texttt{pal43@purdue.edu} \\
}

\begin{document}
\maketitle
\begin{abstract}
This paper studies reinforcement learning for high-frequency trading on limit order books by pairing an Order-Flow-based state model with policy-gradient methods. Instead of value-based RL techniques like tabular Q-learning, our approach deploys policy-based methods like vanilla PPO and DeepSeekMath-inspired variants like GRPO and GSPO, that use group-normalized updates and downside-aware shaping. On backtests with financial assets AMZN, AAPL, and GOOG under a simplified backtesting setup based on spread-scaled rewards, these new policies improve net average PnL, profitability, and drawdown over the Q-Learning baseline. Our results show that (1) Order-Flow signals are an adequate state for policy RL and (2) group-aware PPO surrogates are preferable over value-based baselines. 
\end{abstract}


\def\thefootnote{\dag}\footnotetext{These authors contributed equally to this work.}
\def\thefootnote{\arabic{footnote}}

\section{Introduction}
Directional trading using limit-order-book signals is a central task in quantitative finance. A market facilitates the exchange of assets between buyers and sellers, where profits are made by buying low and selling high (or vice versa through short selling). Orders are matched in a first-in-first-out (FIFO) manner, with limit orders queued until filled by opposing market orders. More than half of modern exchanges now operate electronic limit order books (LOBs). The average of the best bid and ask represents the current market price (also known as mid-price), while the bid-ask spread measures liquidity. Market movements occur when orders are filled or new limit orders are placed within the spread, shifting the best bid or ask levels. An LOB records every resting order; thus, the LOB's dimensionality is enormous, complicating any conditional analysis. Effective modeling, therefore, requires a compact representation that preserves the crucial micro-structural signals \cite{gould2013limit}. Against this backdrop, building an autonomous directional trading agent is both practical and challenging. We emphasize that the environment studied in this paper is a directional trading environment rather than a full market-making environment: the agent selects directional actions from forecast-based states, and does not post bid and ask quotes, manage queue position, or optimize inventory around two-sided liquidity provision.

In this paper, our focus is on forecasting returns across multiple horizons using \textbf{Order-Flow Imbalance}(OFI) features and training three policy-based reinforcement learning models on three financial instruments (AAPL, AMZN, GOOG) that provide trading signals. These new adaptive directional trading agents, trained in simulation includes a deep neural network architecture, serving as the function approximator that extracts features directly from OFI values derived from LOB states. The study addresses three limitations of previous RL based approaches: (i) actor-critic and other methods like PPO are notoriously unstable due to sparse, noisy, and delayed rewards, (ii) financial data is generally non-stationary in practice with high variance in returns and state transitions which is not effectively captured and (iii) it becomes computationally infeasible with the use of critic network for such high-dimensional data like the limit order book. Our main contributions are:

\begin{itemize}
\item First adaptation of DeepSeek-style group-aware RL to finance: To the best of our knowledge, this is among the first applications of DeepSeekMath-inspired Group Relative Policy Optimization (GRPO) and Qwen's Group Sequence Policy Optimization (GSPO) to high-frequency trading, redesigning their group-normalized baselines and sequence-level importance ratio objectives to work with trading episodes and noisy market rewards. While these group-aware objectives have not previously been studied in trading, they have shown strong empirical stability and performance in large language model training \cite{shao2024deepseekmathpushinglimitsmathematical}.
\item  Empirical comparison of group-aware policy optimization in a directional trading setting: Under this limited experimental setup, GRPO and GSPO show higher observed average PnL/profitability and lower observed drawdown than PPO and Q-learning. We present these findings as empirical observations specific to this setup rather than as a general claim of superiority across trading environments. Prior work in this setting \cite{jaddu2023combiningdeeplearningorder} primarily relied on value-based methods, which are often observed to exhibit training instability and comparatively weaker performance under similar market-simulation conditions.
\end{itemize}

Our goal is not to provide a production-ready trading system or a comprehensive market-microstructure evaluation. Rather, we study whether group-aware policy optimization objectives originally developed in LLM training can be adapted to a short-horizon LOB-based directional trading setting and whether they exhibit improved empirical behavior relative to PPO and tabular Q-learning under a common backtesting protocol. Accordingly, the conclusions of this paper should be interpreted within the limits of the chosen assets, sample, reward design, and simplified execution assumptions.

\section{Related Work}

Early directional trading theory begins with \cite{demsetz1968cost}, who viewed the bid-ask spread as compensation for immediacy. \cite{garman1976market} developed an inventory-based MM model, later extended to multiple dealers \cite{stoll1978pricing}. \cite{avellaneda2008high}, following \cite{ho1980dealer}, used stochastic control to derive optimal quotes, assuming continuous prices; \cite{guilbaud2013optimal} adapted the framework to discrete ticks. In all such models, closed-form strategies require restrictive assumptions on price and order arrivals \cite{chan2001electronic}. The advent of millisecond-level LOB data spurred empirical micro-structure studies as in \cite{garman1976market,o1998market}. Reinforcement-learning (RL) methods \cite{zhang2022new,chakrabarty2025time,bhattacharyya2026adaptive,banerjee2026small} have been applied to trading \cite{deng2016deep,jaddu2023combiningdeeplearningorder,wei2019model}, optimal execution \cite{nevmyvaka2006reinforcement}, and portfolio allocation \cite{wang2021deeptrader,jin2016portfolio}. Here, we evaluate two recent group-based policy-optimization algorithms originally proposed to successfully stabilize Large Language Model(LLM) training: \textit{Group Relative Policy Optimization} (GRPO) \cite{shao2024deepseekmathpushinglimitsmathematical} from DeepSeekMath and \textit{Group Sequence Policy Optimization} (GSPO) \cite{gspo} from Qwen (Alibaba Inc.). No recent studies have applied GRPO or GSPO to directional trading or trading-strategy problems. Hence, our goal is to adapt these popular methods to the trading setting and perform an empirical comparison against a standard value-based baseline (tabular Q-Learning). It is also important to distinguish the present setting from classical market-making formulations. Traditional market-making models study two-sided quoting, inventory control, spread placement, and queue dynamics, often under stylized stochastic assumptions. By contrast, the present paper studies a directional trading problem in which the agent acts on forecast-based states extracted from the limit order book.

\section{Methodology}

\subsection{Datasets}
Our experiments use the LOBSTER dataset \cite{Huang2011-tp}, which is derived from the NASDAQ TotalView-ITCH feed and provides high-frequency order-book information together with order-event messages. For each ticker, we use level-10 order-book data and construct the supervised and reinforcement-learning inputs from the resulting event stream. Throughout the paper, the dataset should be interpreted as a limited-sample empirical testbed rather than as a broad multi-regime study. The event counts reported below correspond to the number of usable state transitions after preprocessing.

For each instrument, we use data for the ticker-date pair(s) listed in Table~\ref{tab:data_details}. The train/validation/test partition is applied chronologically at the episode level using an 80\%/10\%/10\% split. The split is chronological, and the forecasting model is trained only on the training portion; validation is used for early stopping, and the test split is held out for final RL evaluation. Each datapoint corresponds to \textit{one trading day} dated June 21, 2013 with timestamps between 9:30 to 10:30 hours having decimal precision of at least milliseconds and up to nanoseconds. Because the evaluation uses three instruments over a single one-hour window from one trading day, the results should be interpreted as a preliminary exploratory study rather than evidence of broad market-regime robustness.

\begin{table}[!ht]
\centering
\begin{tabular}{lccc}
\hline
\textbf{} & \textbf{AMZN} & \textbf{AAPL} & \textbf{GOOG} \\
\hline
Total Data Points & 269,748 & 400,391 & 147,916 \\
\hline
\end{tabular}
\caption{LOBSTER Level-10 datasets for AMZN, AAPL, and GOOG}
\label{tab:data_details}
\end{table}

\subsection{Order Book Feature Extraction}
Let the observable order-book quantities at time $t$ be the prices and aggregated volumes at the first ten non-empty levels: $\{a_t^{i},\,v_t^{i,a},\; b_t^{i},\,v_t^{i,b}\}_{i=1}^{10}$, where $a_t^{i}$ and $b_t^{i}$ are the ask and bid prices at level $i$ and $v_t^{i,a},v_t^{i,b}$ are the aggregated limit order volumes at that level. By following the order-flow extraction method in \cite{jaddu2023combiningdeeplearningorder}, for each level $i$, we compute the \emph{ask order flow} $a\mathrm{OF}_{t,i}$ and \emph{bid order flow} $b\mathrm{OF}_{t,i}$ as the piecewise quantities as shown below:

\begin{align*}
a\mathrm{OF}_{t,i} &=
\begin{cases}
\;v_t^{\,i,a}, & \text{if } a_t^{i} < a_{t-1}^{i},\\[4pt]
\;v_t^{\,i,a}-v_{t-1}^{\,i,a}, & \text{if } a_t^{i} = a_{t-1}^{i},\\[4pt]
\;-\,v_{t-1}^{\,i,a}, & \text{if } a_t^{i} > a_{t-1}^{i},
\end{cases} \quad \\
b\mathrm{OF}_{t,i} &=
\begin{cases}
\;v_t^{\,i,b}, & \text{if } b_t^{i} > b_{t-1}^{i},\\[4pt]
\;v_t^{\,i,b}-v_{t-1}^{\,i,b}, & \text{if } b_t^{i} = b_{t-1}^{i},\\[4pt]
\;-\,v_{t-1}^{\,i,b}, & \text{if } b_t^{i} < b_{t-1}^{i}.
\end{cases}
\label{eq:OF}
\end{align*}

Next, we form the \emph{order-flow imbalance} (OFI) per level as the bid minus ask flows $\mathrm{OFI}_t \;=\; b\mathrm{OF}_t - a\mathrm{OF}_t \in\mathbb{R}^{10}$. Our pipeline uses a supervised feature vector $x_t$ defined as $x_t \;=\; \mathrm{OFI}_t$ optionally normalized per sample by its maximum absolute component: $x_t \leftarrow \frac{\mathrm{OFI}_t}{\max\bigl(1,\;\|\mathrm{OFI}_t\|_\infty\bigr)}$. 

Thus, each learning instance is anchored at an event index $t$ and uses a ten-dimensional OFI vector derived from the first ten non-empty book levels. The optional normalization above is applied elementwise through division by the maximum absolute component of the OFI vector so that unusually large instantaneous imbalances do not dominate the scale of the state representation. Unless otherwise stated, time indices in the sequel refer to decision steps aligned with the event stream after preprocessing.

\begin{figure*}[ht]
    \centering
    \begin{subfigure}[b]{0.23\textwidth}
        \centering
        \includegraphics[trim={0 0 0 0.9cm},clip,width=\textwidth]{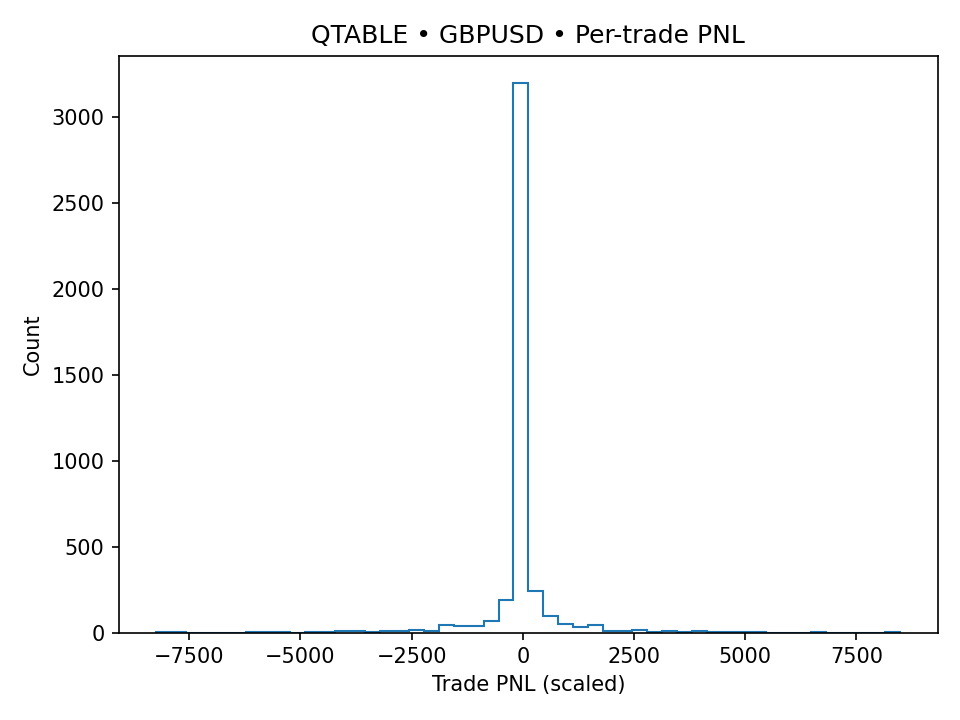}
        \caption{QTable - Histogram}
        \label{fig:goog_1a}
    \end{subfigure}
    \begin{subfigure}[b]{0.23\textwidth}
        \centering
        \includegraphics[trim={0 0 0 0.9cm},clip,width=\textwidth]{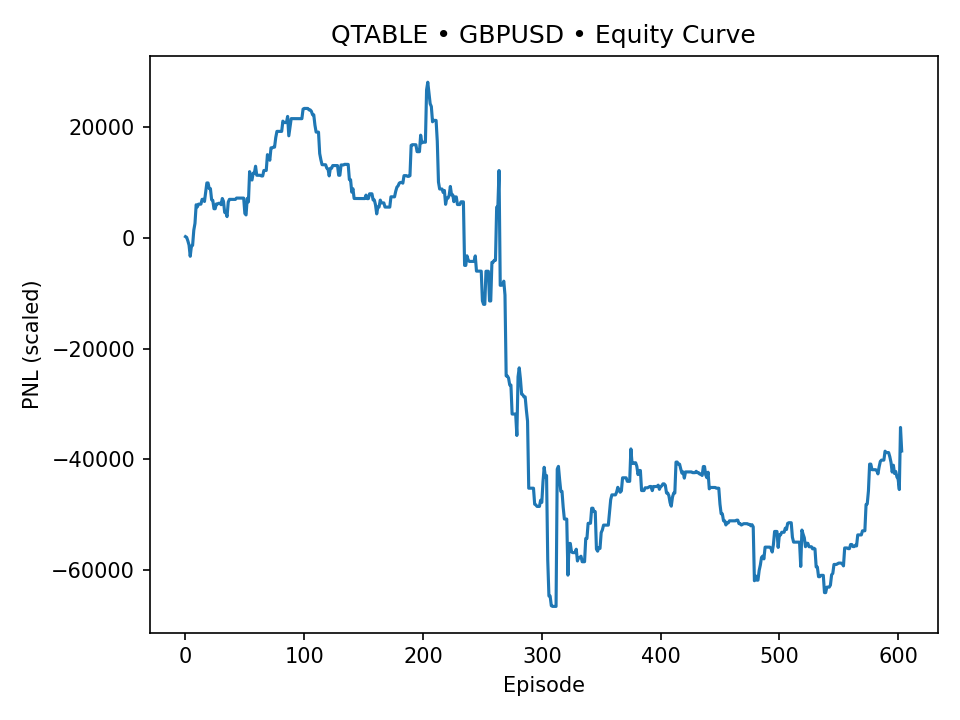}
        \caption{QTable - Equity}
        \label{fig:goog_1b}
    \end{subfigure}
    \begin{subfigure}[b]{0.23\textwidth}
        \centering
        \includegraphics[trim={0 0 0 0.9cm},clip,width=\textwidth]{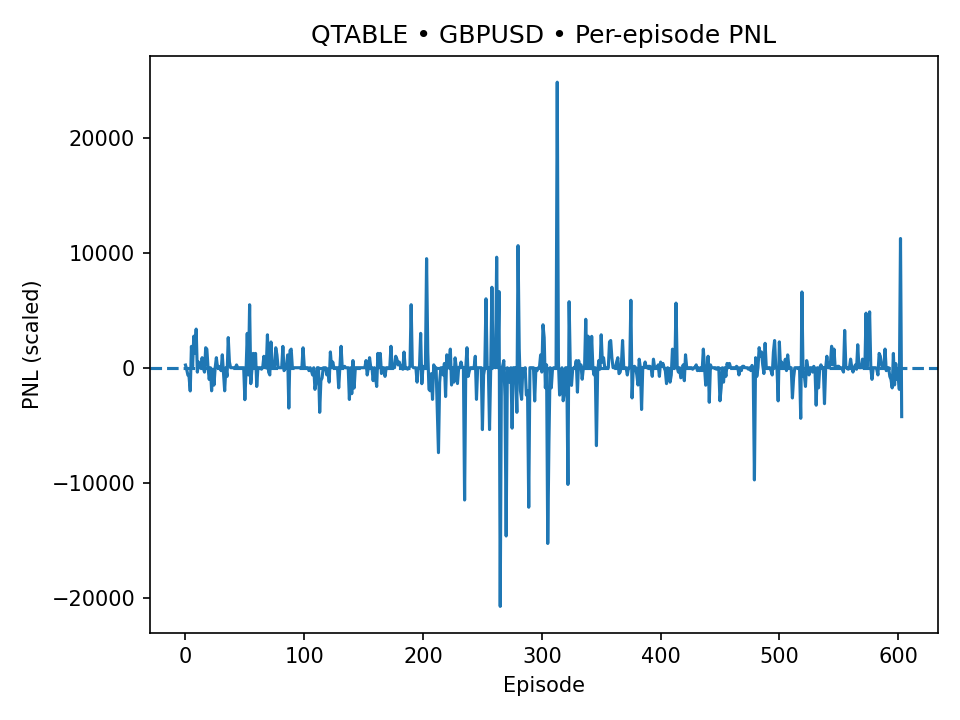}
        \caption{QTable - Returns}
        \label{fig:goog_1c}
    \end{subfigure}
    \begin{subfigure}[b]{0.23\textwidth}
        \centering
        \includegraphics[trim={0 0 0 0.9cm},clip,width=\textwidth]{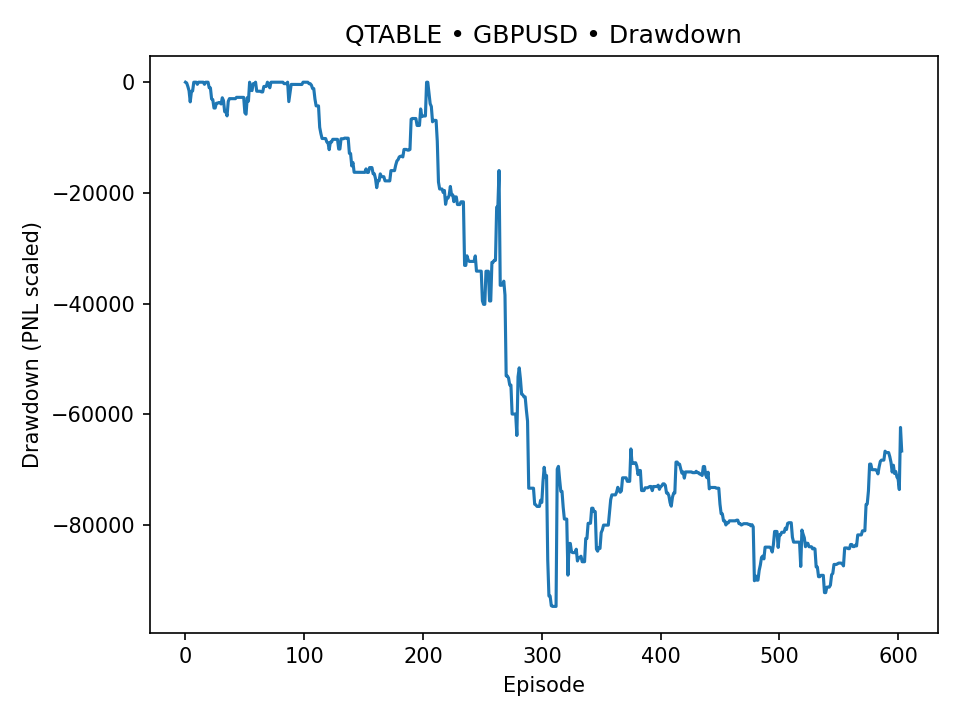}
        \caption{QTable - Drawdown}
        \label{fig:goog_1d}
    \end{subfigure}
    \\[0.8em]

    \begin{subfigure}[b]{0.23\textwidth}
        \centering
        \includegraphics[trim={0 0 0 0.9cm},clip,width=\textwidth]{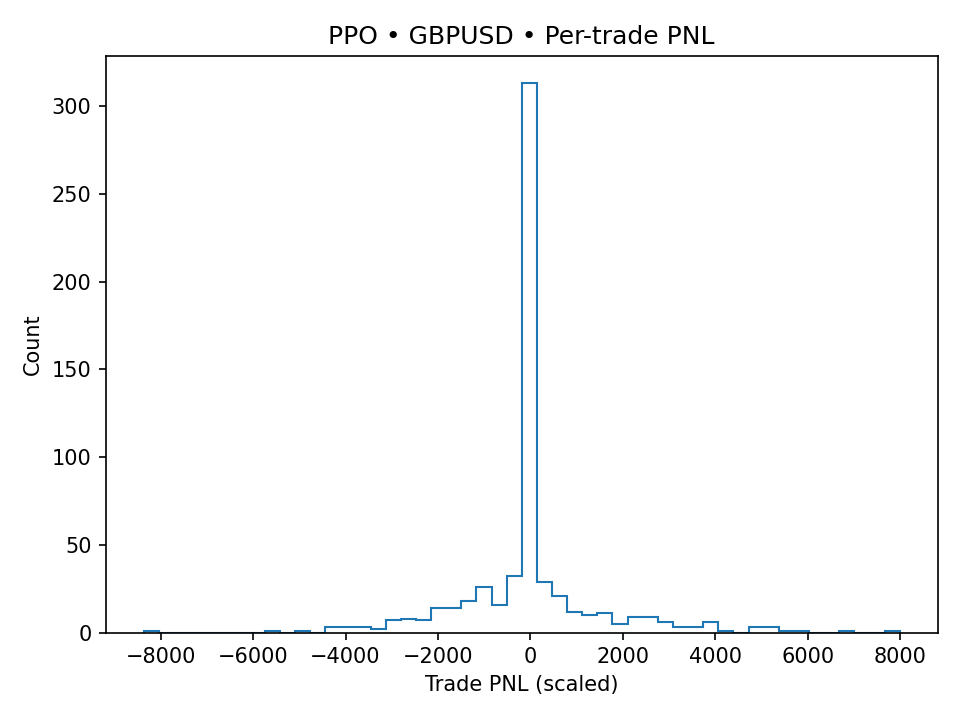}
        \caption{PPO - Histogram}
        \label{fig:goog_1e}
    \end{subfigure}
    \begin{subfigure}[b]{0.23\textwidth}
        \centering
        \includegraphics[trim={0 0 0 0.9cm},clip,width=\textwidth]{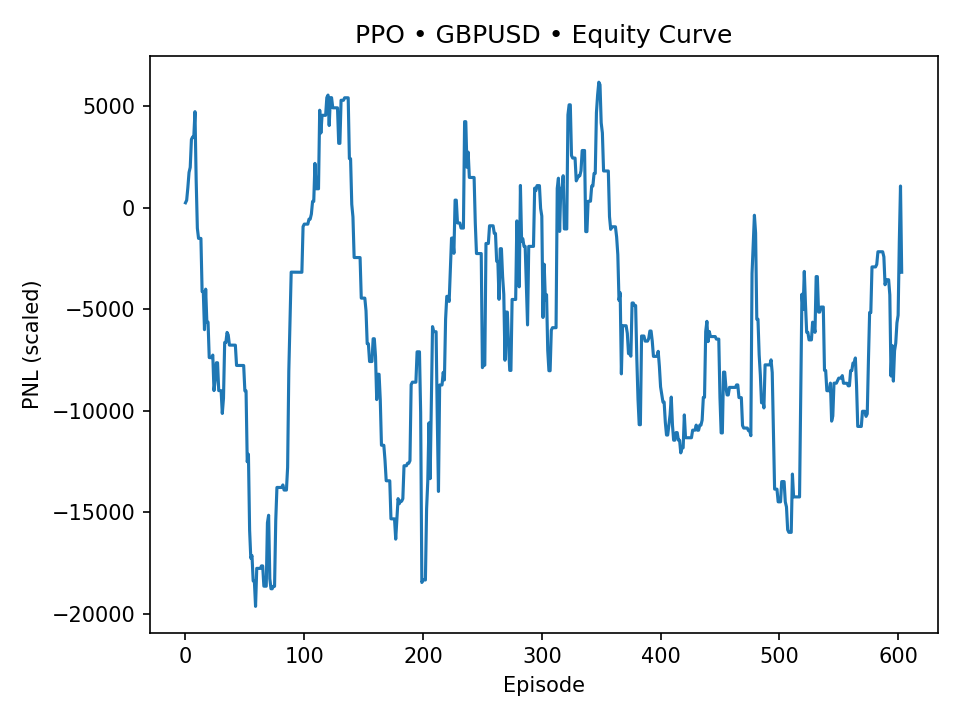}
        \caption{PPO - Equity}
        \label{fig:goog_1f}
    \end{subfigure}
    \begin{subfigure}[b]{0.23\textwidth}
        \centering
        \includegraphics[trim={0 0 0 0.9cm},clip,width=\textwidth]{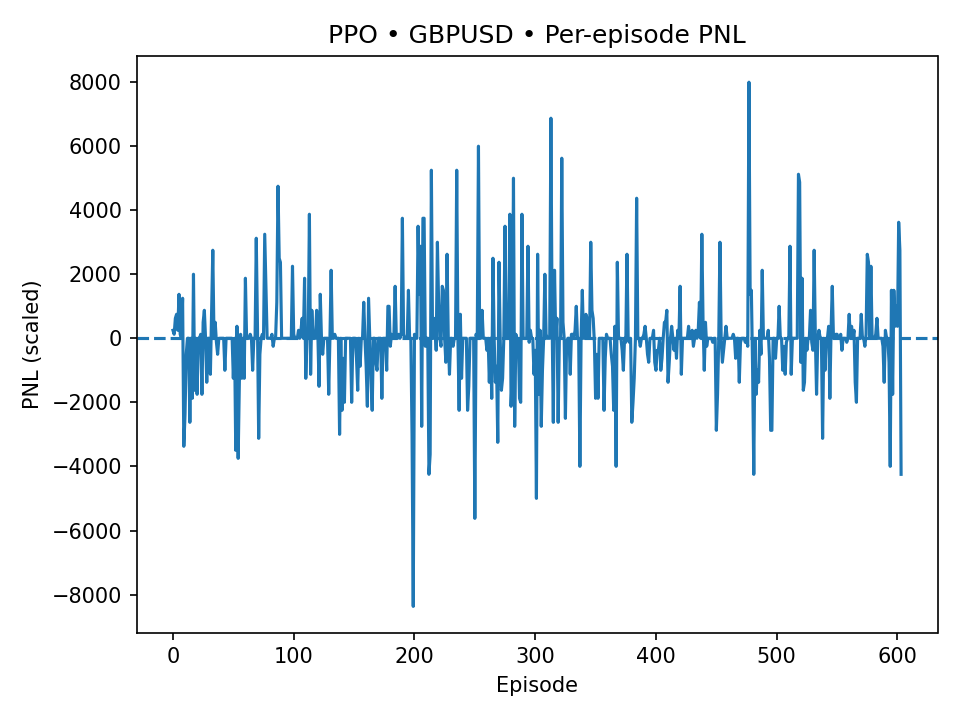}
        \caption{PPO - Returns}
        \label{fig:goog_1g}
    \end{subfigure}
    \begin{subfigure}[b]{0.23\textwidth}
        \centering
        \includegraphics[trim={0 0 0 0.9cm},clip,width=\textwidth]{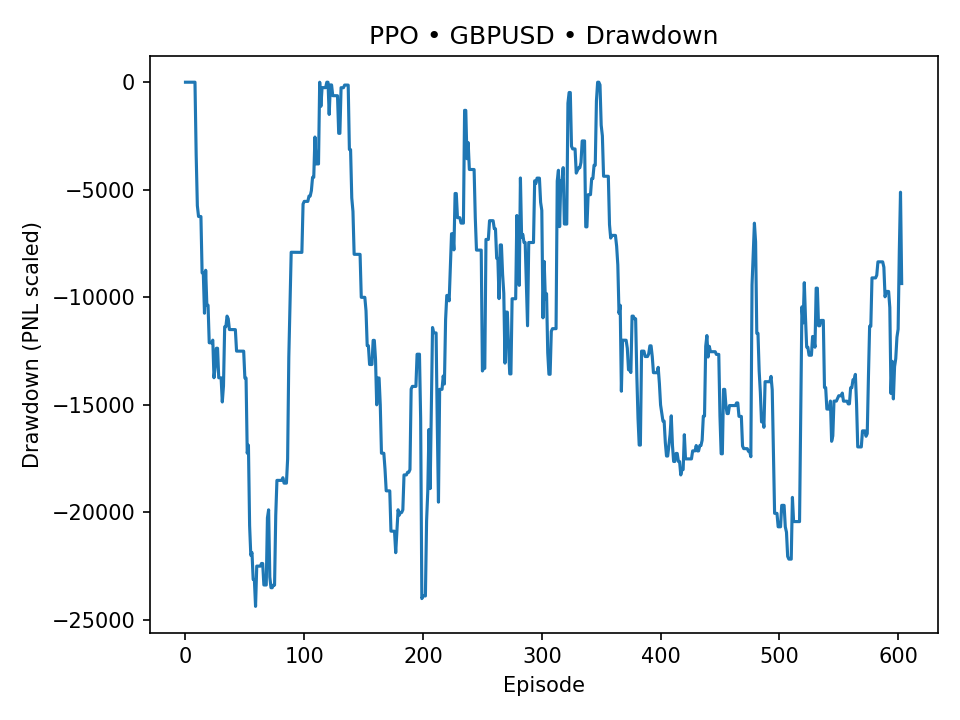}
        \caption{PPO - Drawdown}
        \label{fig:goog_1h}
    \end{subfigure}
    \\[0.8em]

    \begin{subfigure}[b]{0.23\textwidth}
        \centering
        \includegraphics[trim={0 0 0 0.9cm},clip,width=\textwidth]{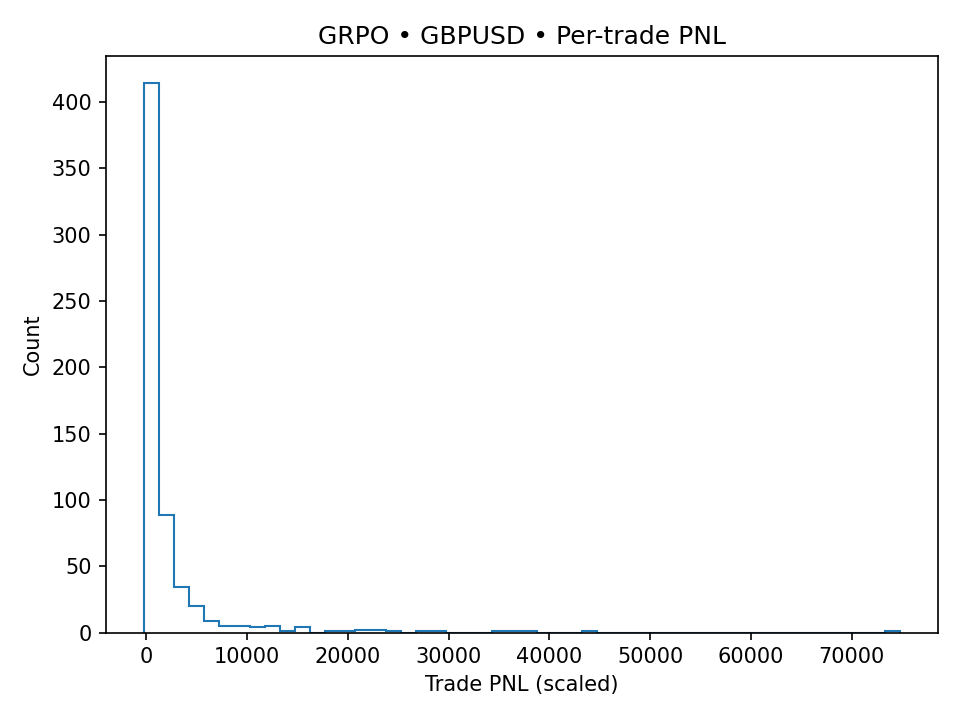}
        \caption{GRPO - Histogram}
        \label{fig:goog_1i}
    \end{subfigure}
    \begin{subfigure}[b]{0.23\textwidth}
        \centering
        \includegraphics[trim={0 0 0 0.9cm},clip,width=\textwidth]{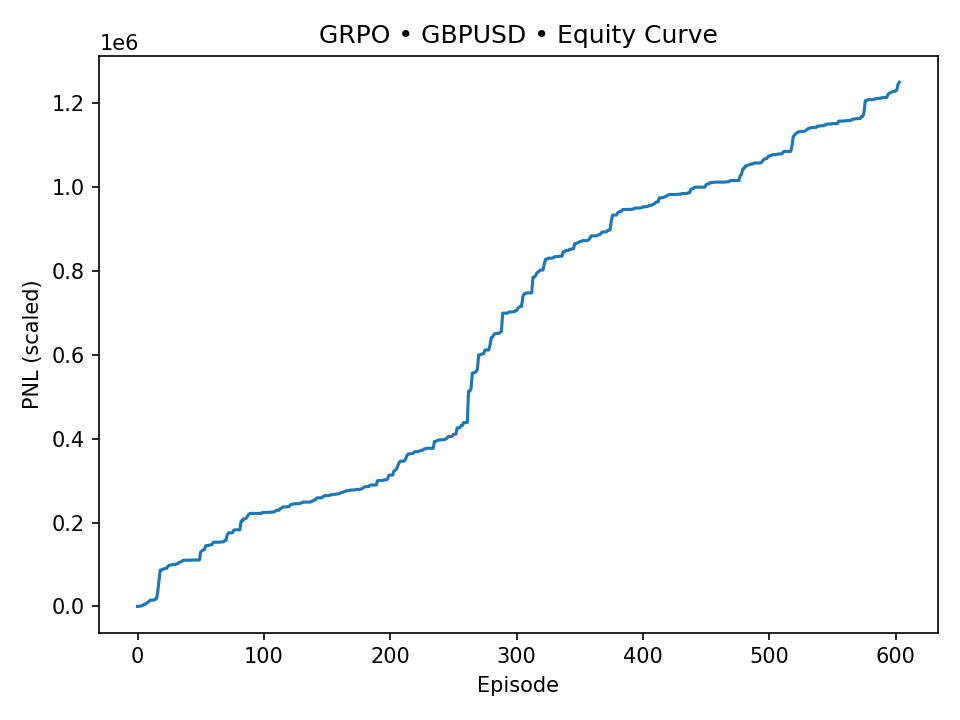}
        \caption{GRPO - Equity}
        \label{fig:goog_1j}
    \end{subfigure}
    \begin{subfigure}[b]{0.23\textwidth}
        \centering
        \includegraphics[trim={0 0 0 0.9cm},clip,width=\textwidth]{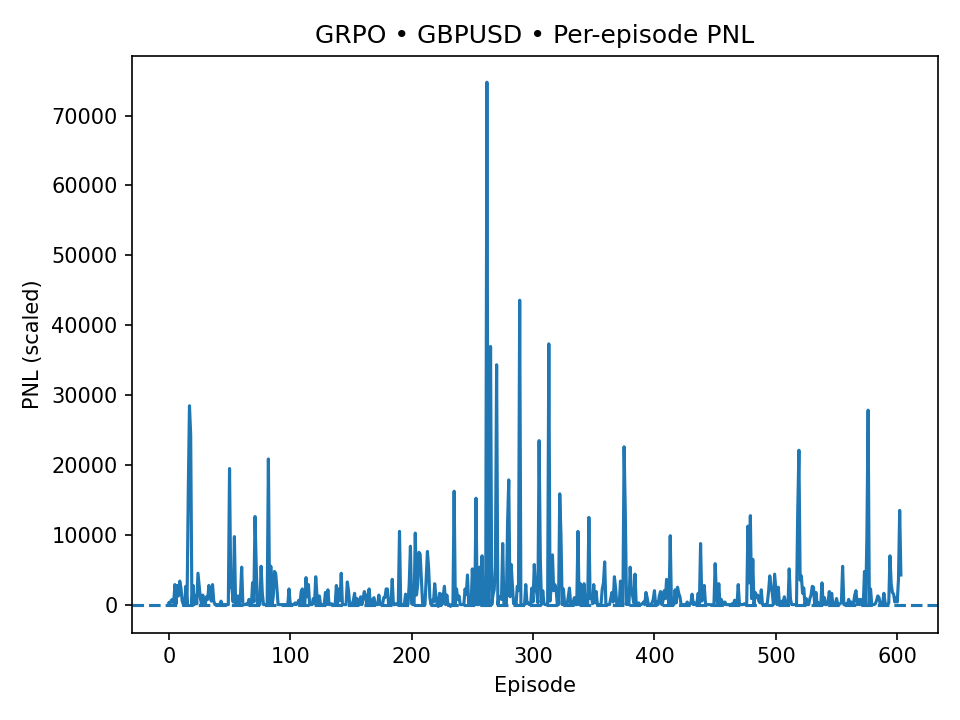}
        \caption{GRPO - Returns}
        \label{fig:goog_1k}
    \end{subfigure}
    \begin{subfigure}[b]{0.23\textwidth}
        \centering
        \includegraphics[trim={0 0 0 0.9cm},clip,width=\textwidth]{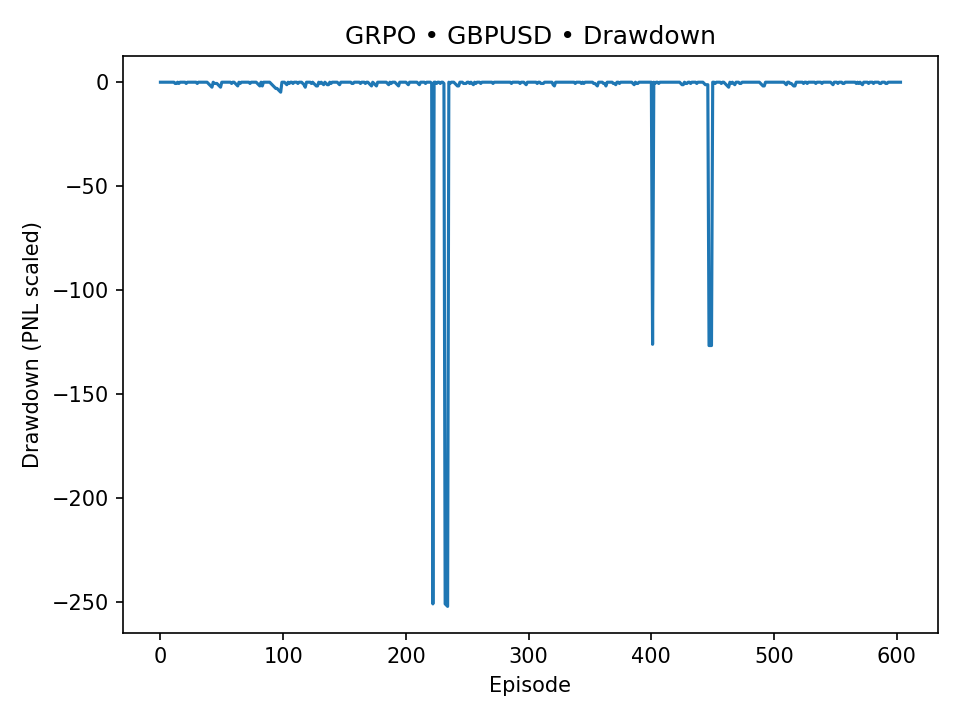}
        \caption{GRPO - Drawdown}
        \label{fig:goog_1l}
    \end{subfigure}
    \\[0.8em]

    \begin{subfigure}[b]{0.23\textwidth}
        \centering
        \includegraphics[trim={0 0 0 0.9cm},clip,width=\textwidth]{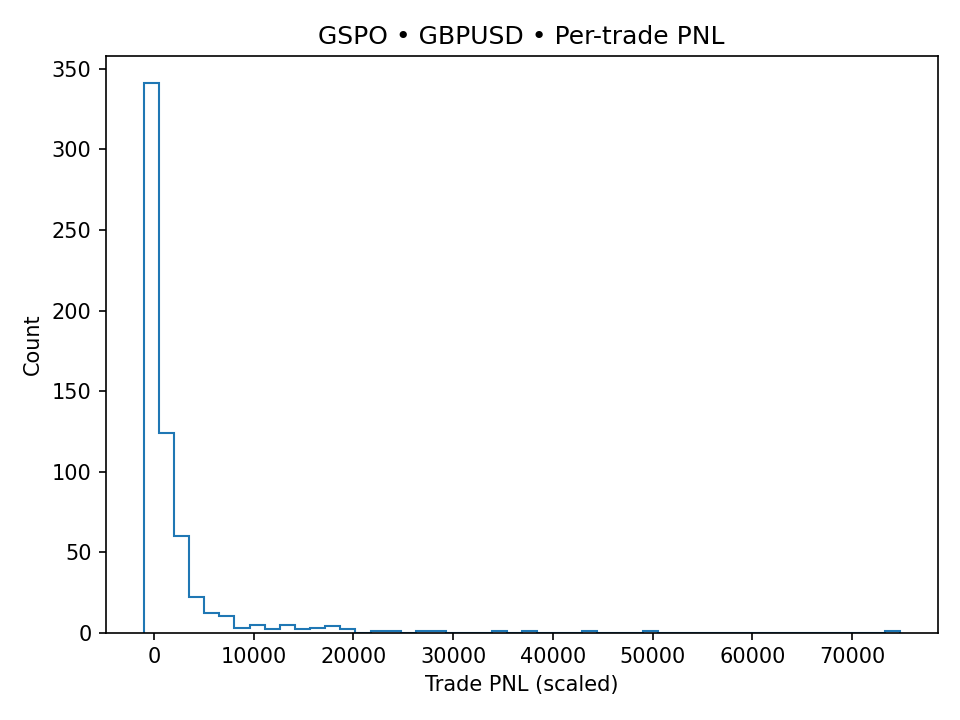}
        \caption{GSPO - Histogram}
        \label{fig:goog_1m}
    \end{subfigure}
    \begin{subfigure}[b]{0.23\textwidth}
        \centering
        \includegraphics[trim={0 0 0 0.9cm},clip,width=\textwidth]{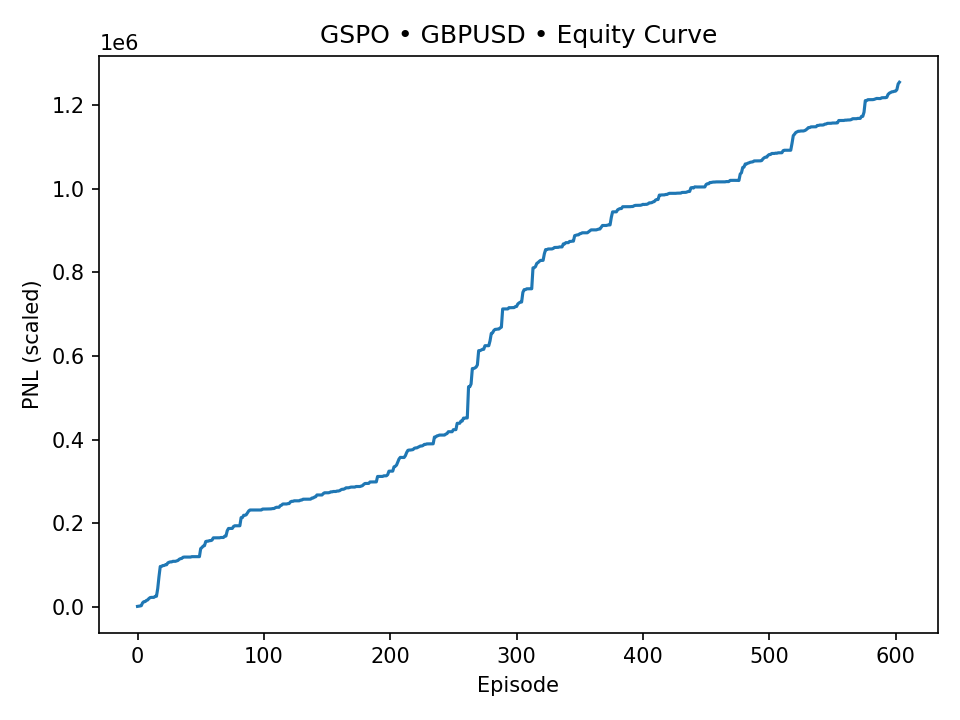}
        \caption{GSPO - Equity}
        \label{fig:goog_1n}
    \end{subfigure}
    \begin{subfigure}[b]{0.23\textwidth}
        \centering
        \includegraphics[trim={0 0 0 0.9cm},clip,width=\textwidth]{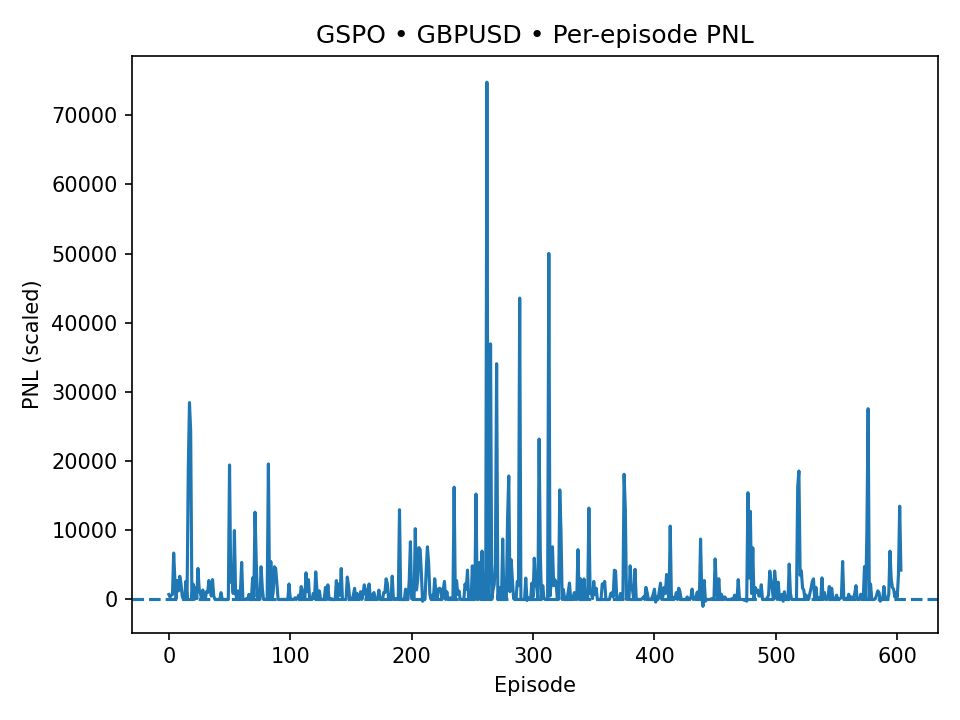}
        \caption{GSPO - Returns}
        \label{fig:goog_1o}
    \end{subfigure}
    \begin{subfigure}[b]{0.23\textwidth}
        \centering
        \includegraphics[trim={0 0 0 0.9cm},clip,width=\textwidth]{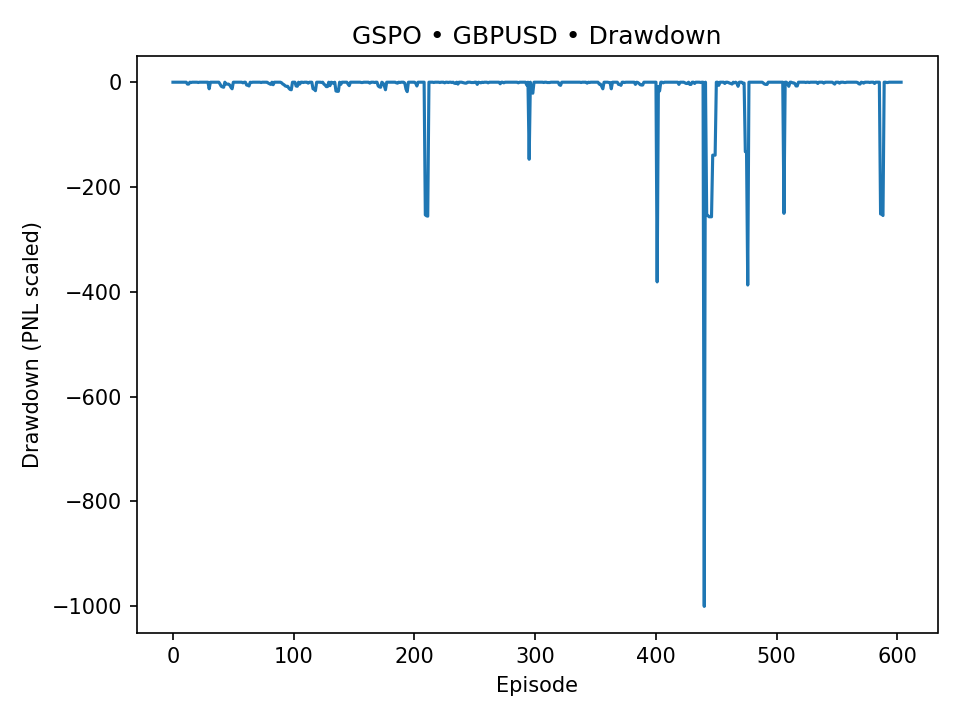}
        \caption{GSPO - Drawdown}
        \label{fig:goog_1p}
    \end{subfigure}

    \caption{Held-out evaluation plots for GOOG showing the trade-PnL histogram, equity curve (scaled by $10^6$), episode-level PnL returns, and drawdown under the backtesting protocol}
    \label{fig:goog_4x4}
\end{figure*}

\subsection{State-Transition Model}
A regression network $f_\theta$ maps this $10$-dimensional OFI vector to $H$ horizon alphas $\hat\alpha_t=f_\theta(x_t)$, which are then concatenated with the previous discrete action to form the state $s_t=[\hat\alpha_t;\,a_{t-1}]\in\mathbb{R}^{H+1}$ for the MDP $\mathcal{M} = (\mathcal{S},\mathcal{A},\mathcal{R},\Omega)$ used by the RL agents. The extractor $f_{\theta}$ is an multilayer perceptron(MLP) model with 4 hidden layers of dimension 2048 trained by minimizing mean squared error(MSE). Optimization uses ADAM with per-instrument learning rates, weight decay, and early stopping on validation loss. The trained $f_{\theta}$ supplies $\hat\alpha_t$ to downstream agents, which are concatenated to represent the state with horizon value $H = 6$. The action space consists of discrete directional trading actions. Reward per episode is computed based on the change in mid-price after the action is taken, scaled by a factor depending on the bid-ask spread.

Let $m_t$ denote the mid-price at decision step $t$, defined from the best bid and ask prices as
\[
m_t = \frac{a_t^{1}+b_t^{1}}{2}.
\]
The supervised forecasting module predicts returns over a fixed set of horizons
\[
\mathcal{H} = \{h_1,\dots,h_H\}.
\]
For each horizon $h \in \mathcal{H}$, we define the target return as
\[
y_t^{(h)} = \frac{m_{t+h}-m_t}{m_t},
\]
where the horizon is measured in event steps/message arrivals in the preprocessed LOBSTER event stream. The forecasting network $f_\theta$ maps the OFI feature vector $x_t \in \mathbb{R}^{10}$ to the $H$-dimensional vector of predicted returns
\[
\hat{\alpha}_t = f_\theta(x_t)
=
\bigl(\hat{y}_t^{(h_1)},\dots,\hat{y}_t^{(h_H)}\bigr)\in \mathbb{R}^{H}.
\]

The forecaster is trained by minimizing mean squared error over all available training indices and horizons:
\[
\mathcal{L}_{\mathrm{sup}}(\theta)
=
\frac{1}{N}
\sum_{t=1}^{N}
\sum_{h \in \mathcal{H}}
\bigl(\hat{y}_t^{(h)}-y_t^{(h)}\bigr)^2.
\]
In our implementation, $f_\theta$ is a multi-layer perceptron with four hidden layers of width 2048. Optimization uses Adam with instrument-specific learning rates, weight decay, and early stopping based on validation loss.

After training, the forecasting network is frozen and used only as a state-construction module for reinforcement learning. The RL state at time $t$ is defined as
\[
s_t = [\hat{\alpha}_t;\, a_{t-1}] \in \mathbb{R}^{H+1},
\]
that is, the concatenation of the multi-horizon predicted return vector and the previous discrete action. This construction yields a compact state representation that summarizes short-horizon directional information from the order book while preserving the agent's most recent trading decision. The paper therefore evaluates RL methods on top of a fixed learned representation rather than jointly optimizing forecasting and control.


\begin{figure*}[ht]
    \centering
    \begin{subfigure}[b]{0.23\textwidth}
        \centering
        \includegraphics[width=\textwidth]{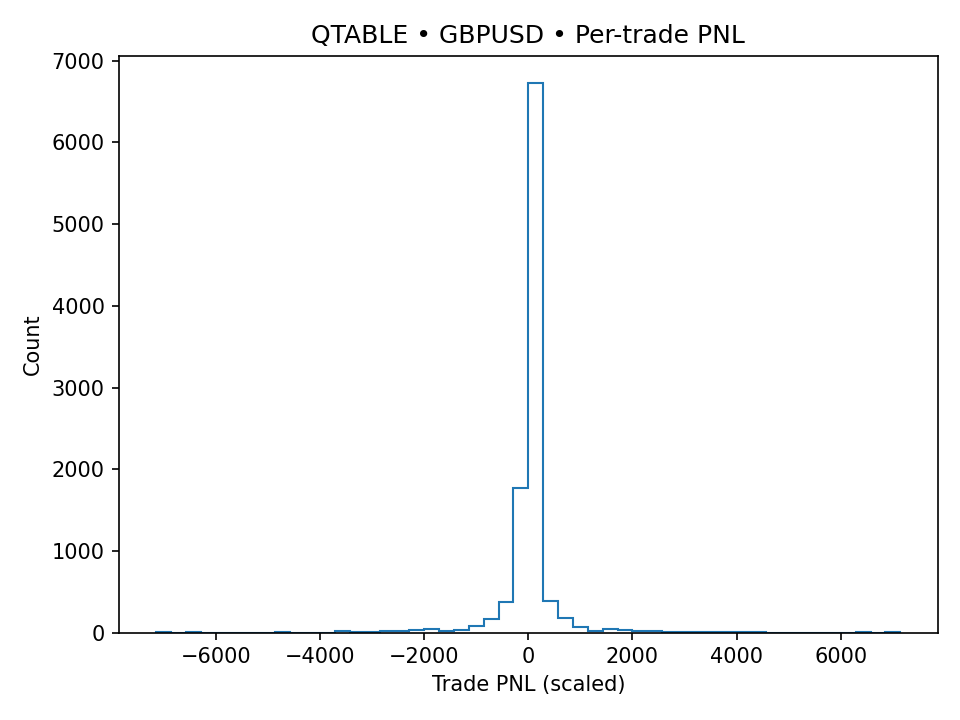}
        \caption{QTable - Histogram}
        \label{fig:amzn_1a}
    \end{subfigure}
    \begin{subfigure}[b]{0.23\textwidth}
        \centering
        \includegraphics[width=\textwidth]{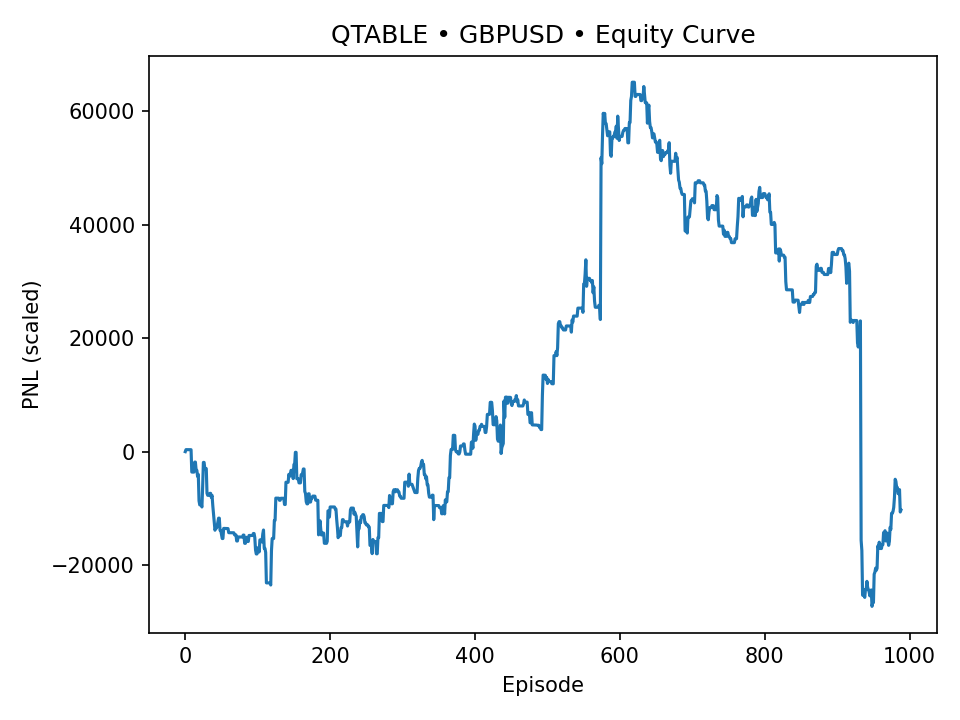}
        \caption{QTable - Equity}
        \label{fig:amzn_1b}
    \end{subfigure}
    \begin{subfigure}[b]{0.23\textwidth}
        \centering
        \includegraphics[width=\textwidth]{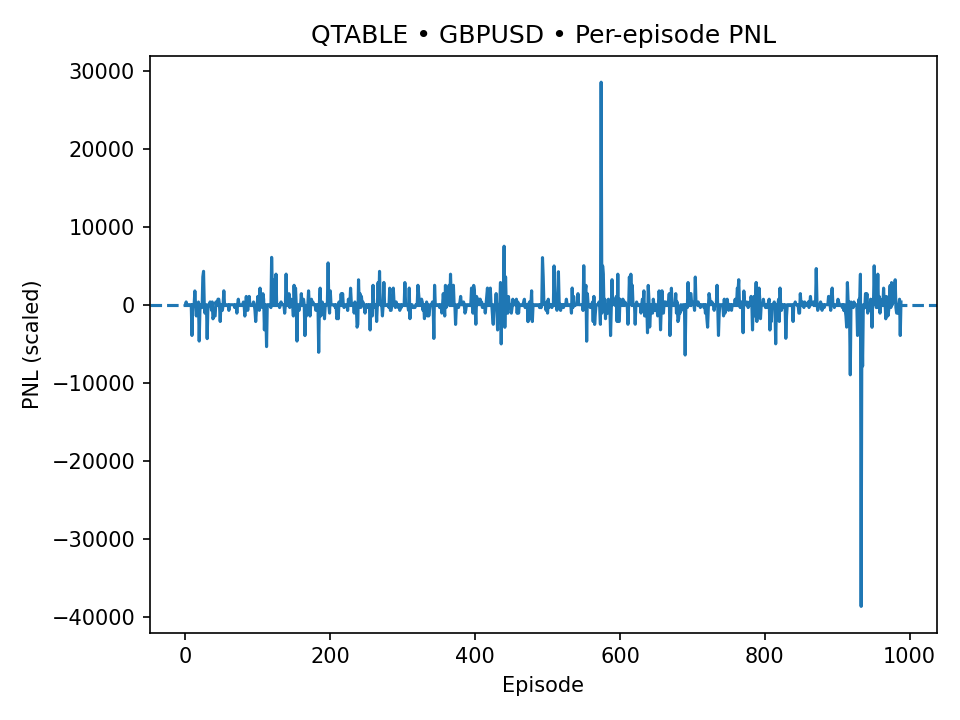}
        \caption{QTable - Returns}
        \label{fig:amzn_1c}
    \end{subfigure}
    \begin{subfigure}[b]{0.23\textwidth}
        \centering
        \includegraphics[width=\textwidth]{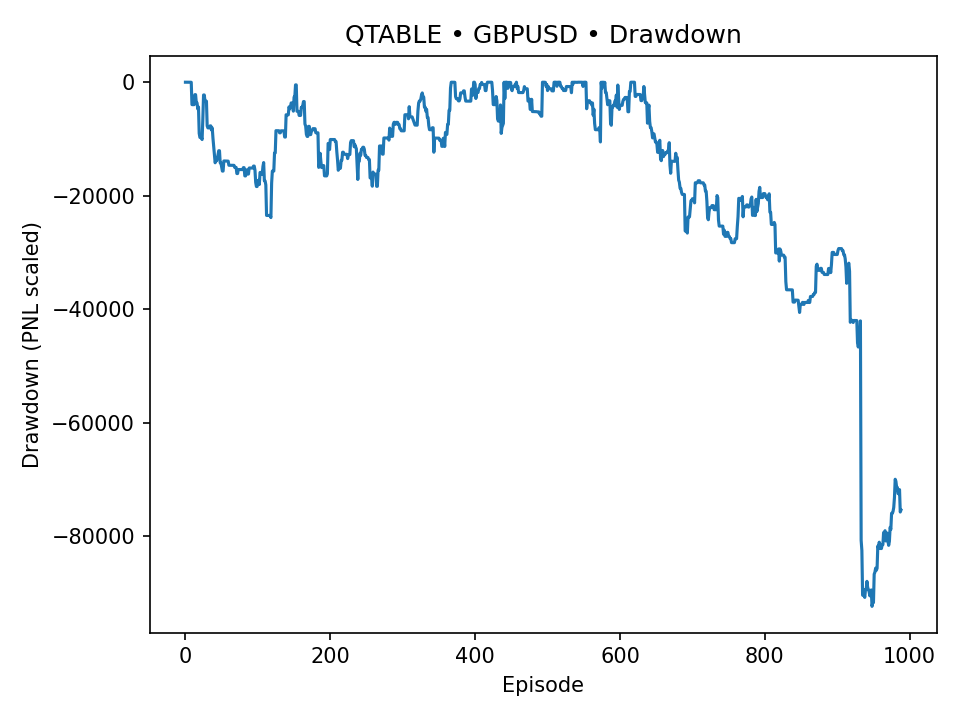}
        \caption{QTable - Drawdown}
        \label{fig:amzn_1d}
    \end{subfigure}
    \\[0.8em]

    \begin{subfigure}[b]{0.23\textwidth}
        \centering
        \includegraphics[width=\textwidth]{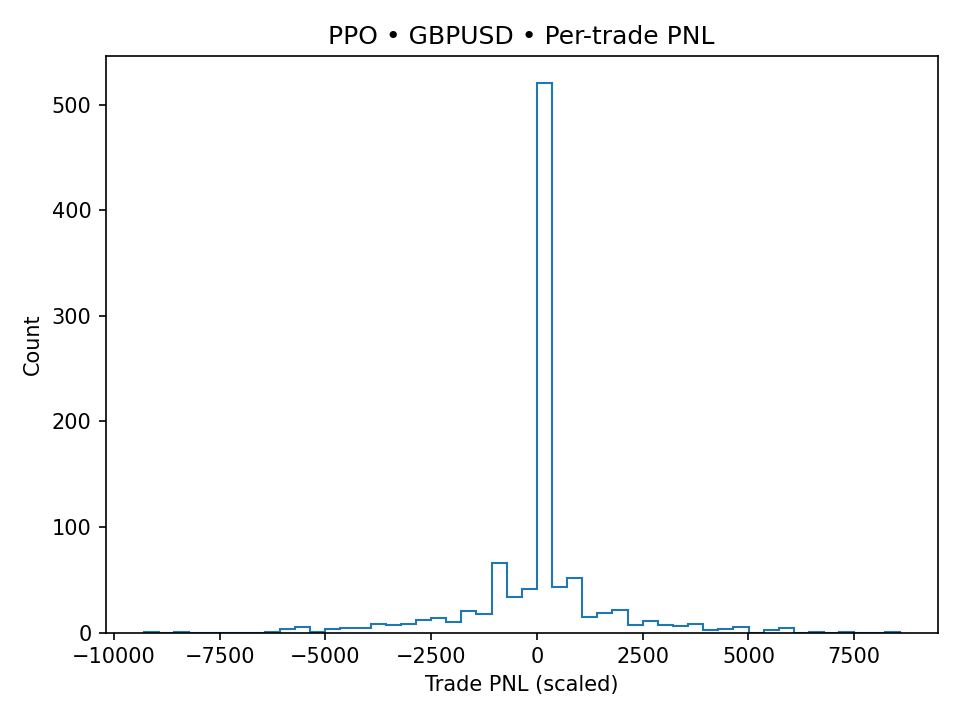}
        \caption{PPO - Histogram}
        \label{fig:amzn_1e}
    \end{subfigure}
    \begin{subfigure}[b]{0.23\textwidth}
        \centering
        \includegraphics[width=\textwidth]{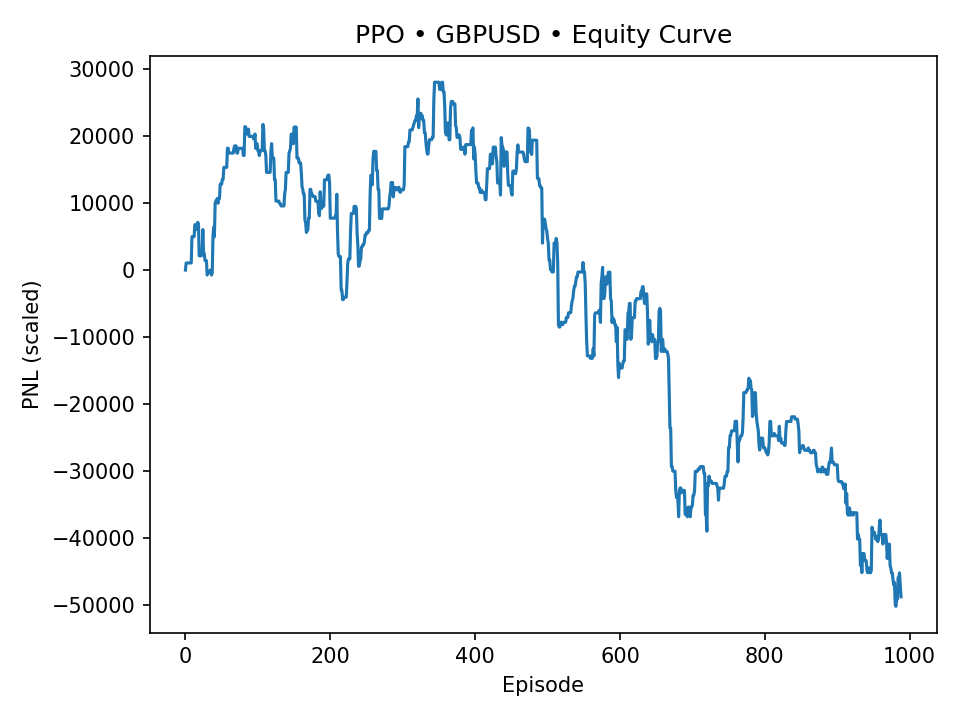}
        \caption{PPO - Equity}
        \label{fig:amzn_1f}
    \end{subfigure}
    \begin{subfigure}[b]{0.23\textwidth}
        \centering
        \includegraphics[width=\textwidth]{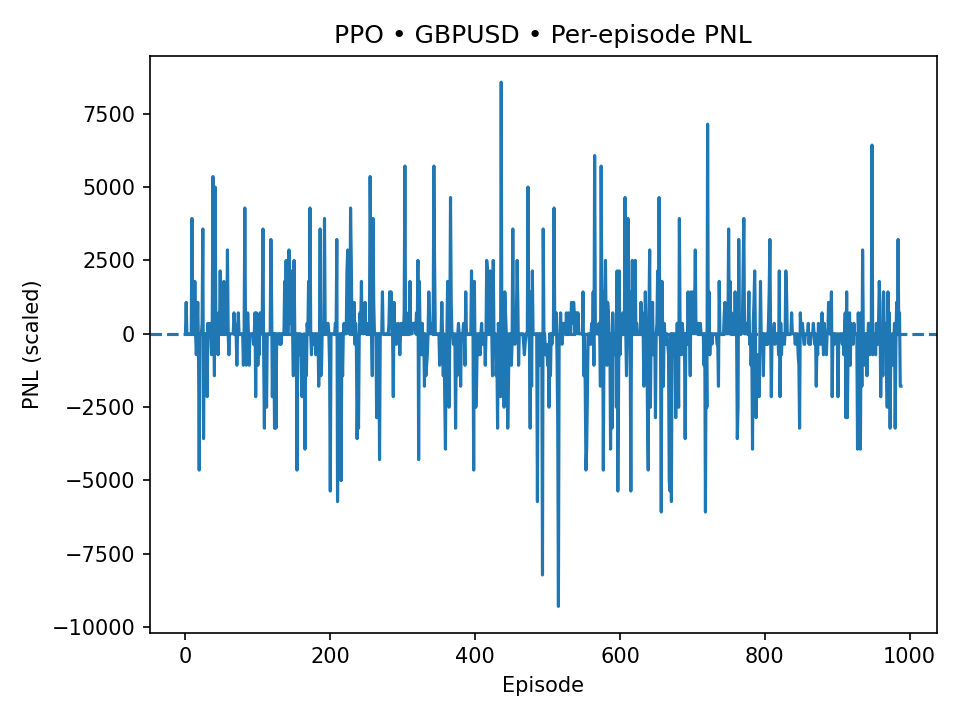}
        \caption{PPO - Returns}
        \label{fig:amzn_1g}
    \end{subfigure}
    \begin{subfigure}[b]{0.23\textwidth}
        \centering
        \includegraphics[width=\textwidth]{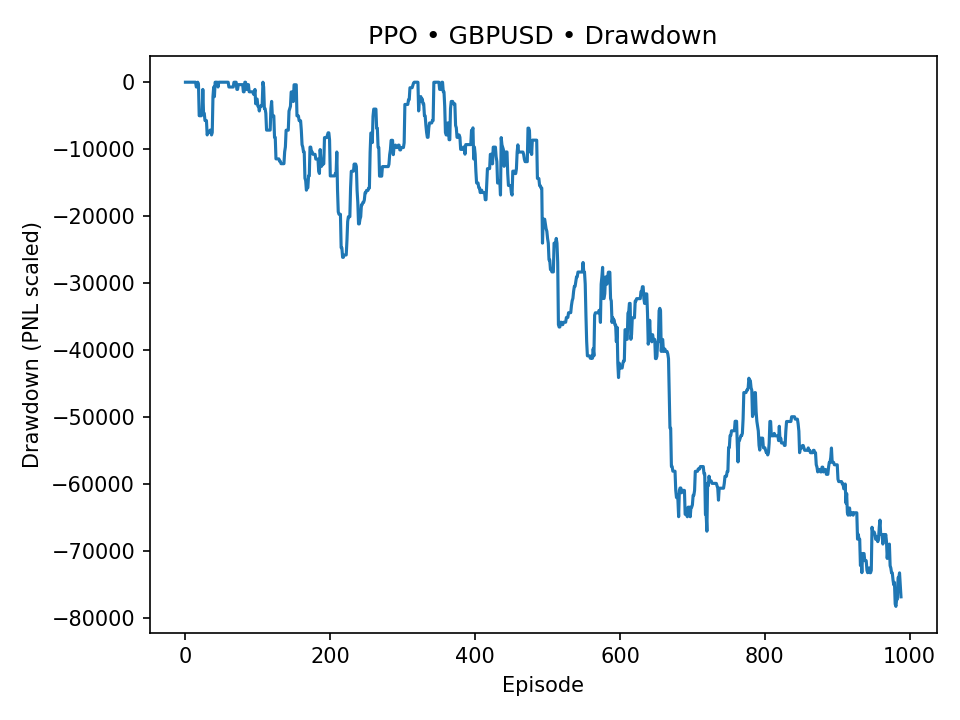}
        \caption{PPO - Drawdown}
        \label{fig:amzn_1h}
    \end{subfigure}
    \\[0.8em]

    \begin{subfigure}[b]{0.23\textwidth}
        \centering
        \includegraphics[width=\textwidth]{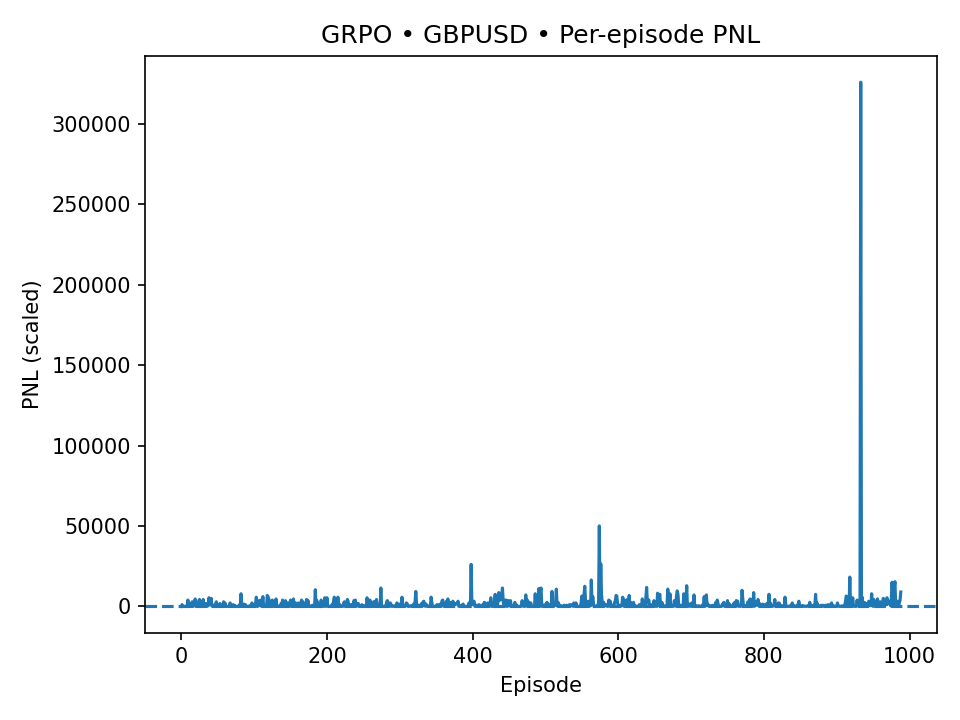}
        \caption{GRPO - Histogram}
        \label{fig:amzn_1i}
    \end{subfigure}
    \begin{subfigure}[b]{0.23\textwidth}
        \centering
        \includegraphics[width=\textwidth]{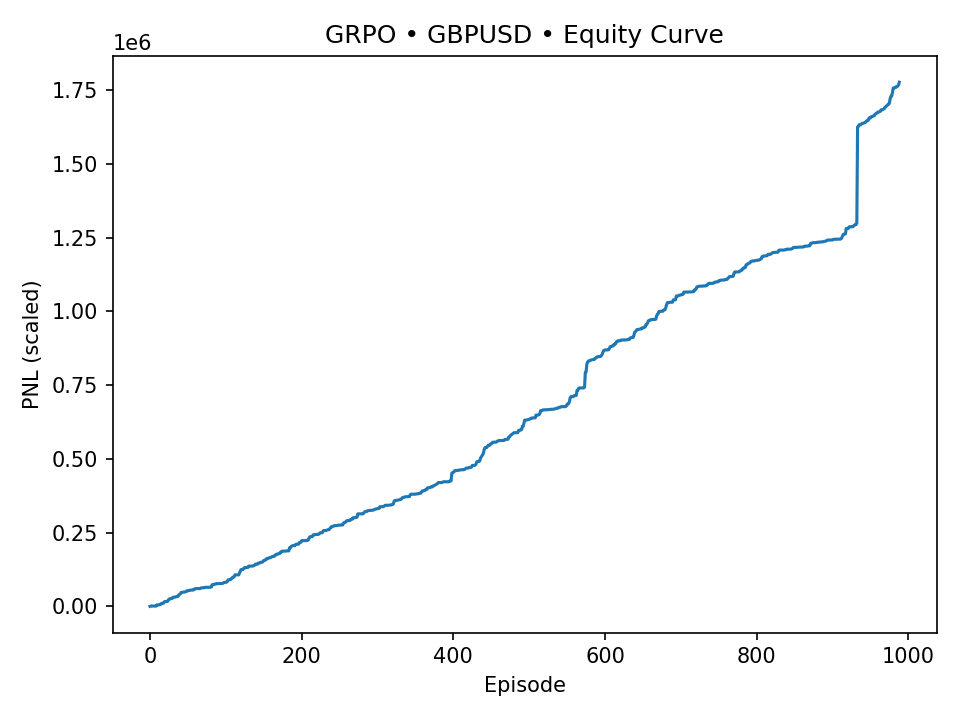}
        \caption{GRPO - Equity}
        \label{fig:amzn_1j}
    \end{subfigure}
    \begin{subfigure}[b]{0.23\textwidth}
        \centering
        \includegraphics[width=\textwidth]{figures/amaz-plots/GBPUSD_grpo_episode_returns.png}
        \caption{GRPO - Returns}
        \label{fig:amzn_1k}
    \end{subfigure}
    \begin{subfigure}[b]{0.23\textwidth}
        \centering
        \includegraphics[width=\textwidth]{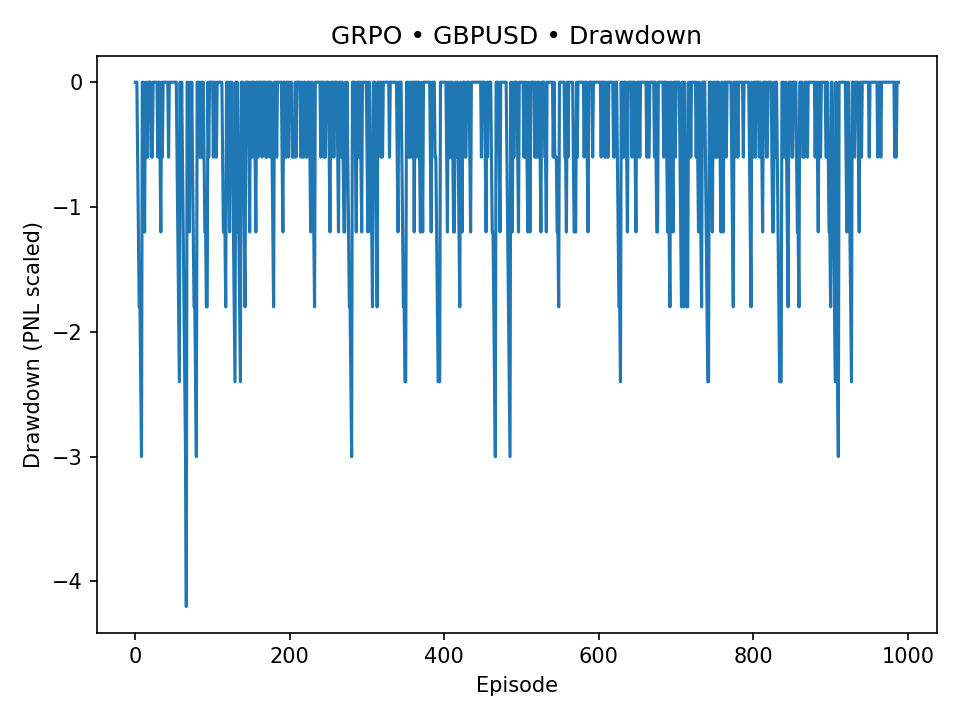}
        \caption{GRPO - Drawdown}
        \label{fig:amzn_1l}
    \end{subfigure}
    \\[0.8em]

    \begin{subfigure}[b]{0.23\textwidth}
        \centering
        \includegraphics[width=\textwidth]{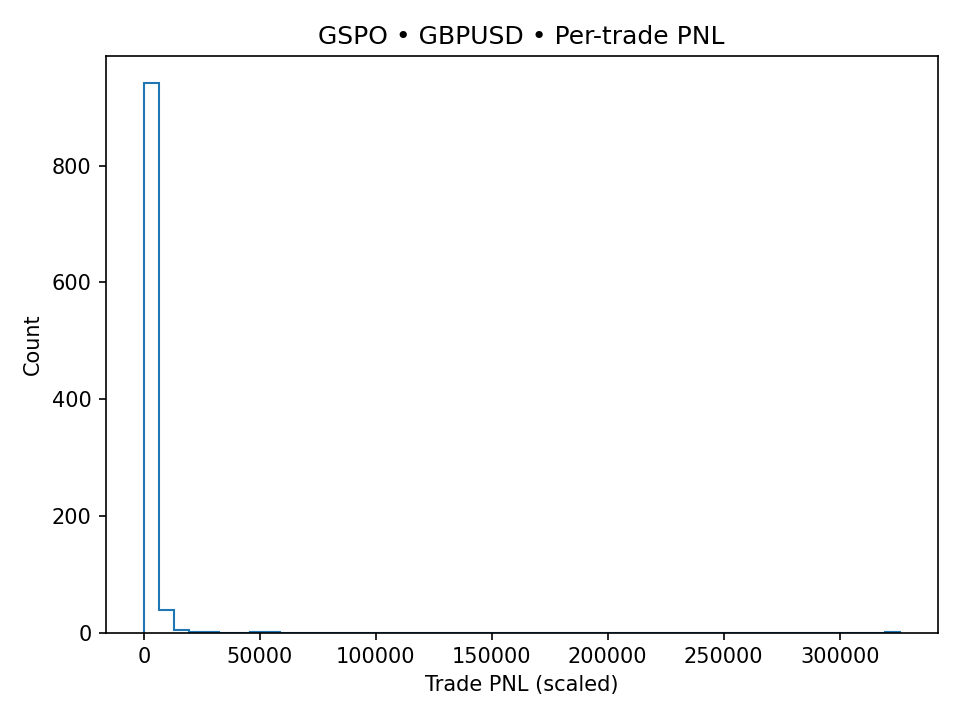}
        \caption{GSPO - Histogram}
        \label{fig:amzn_1m}
    \end{subfigure}
    \begin{subfigure}[b]{0.23\textwidth}
        \centering
        \includegraphics[width=\textwidth]{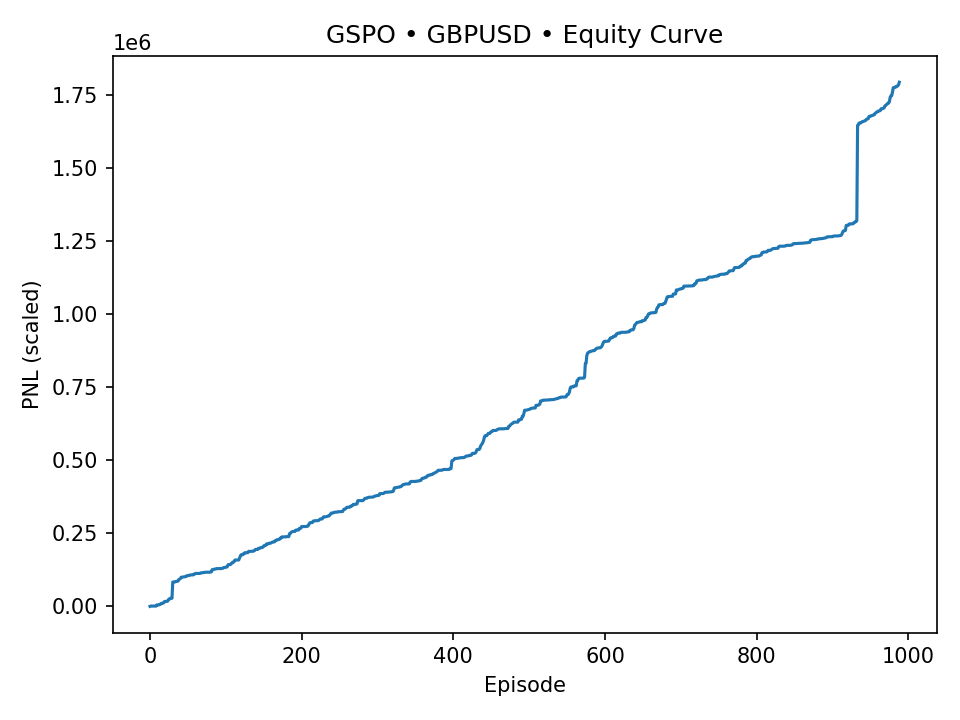}
        \caption{GSPO - Equity}
        \label{fig:amzn_1n}
    \end{subfigure}
    \begin{subfigure}[b]{0.23\textwidth}
        \centering
        \includegraphics[width=\textwidth]{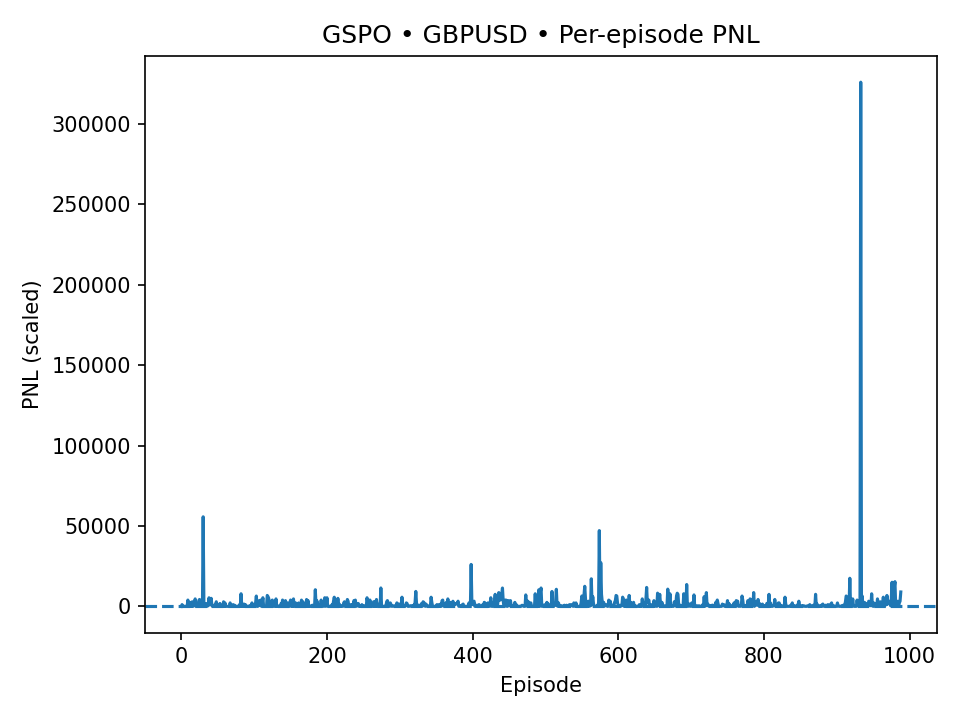}
        \caption{GSPO - Returns}
        \label{fig:amzn_1o}
    \end{subfigure}
    \begin{subfigure}[b]{0.23\textwidth}
        \centering
        \includegraphics[width=\textwidth]{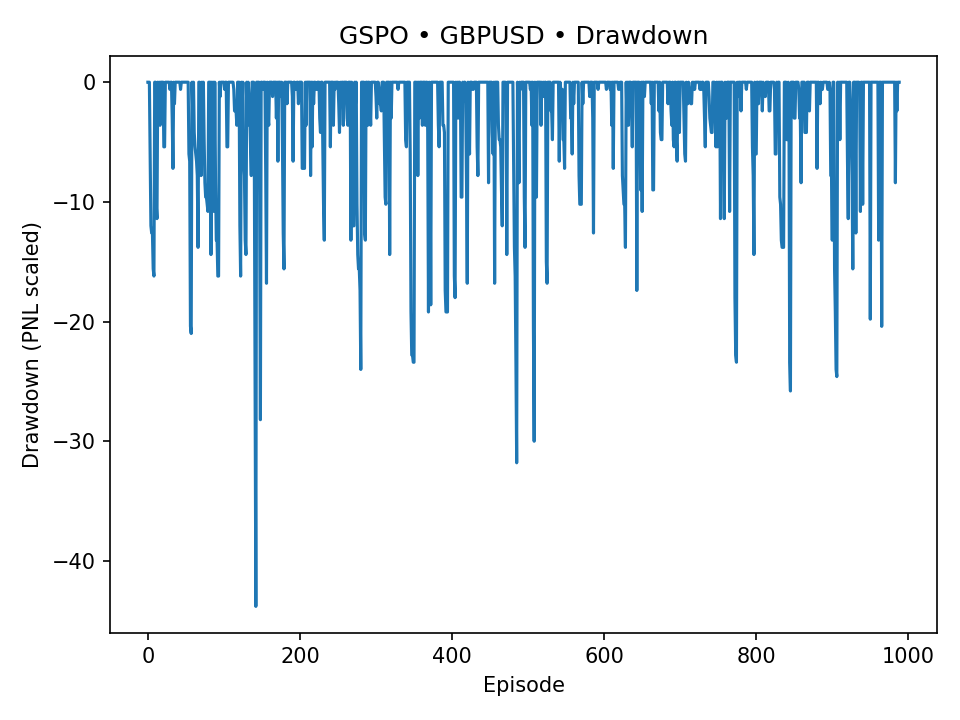}
        \caption{GSPO - Drawdown}
        \label{fig:amzn_1p}
    \end{subfigure}

    \caption{Held-out evaluation plots for AMZN showing the trade-PnL histogram, equity curve (scaled by $10^6$), episode-level PnL returns, and drawdown under the backtesting protocol described in Section~2.}
    \label{fig:amzn_4x4}
\end{figure*}


\begin{figure*}[ht]
    \centering
    \begin{subfigure}[b]{0.23\textwidth}
        \centering
        \includegraphics[width=\textwidth]{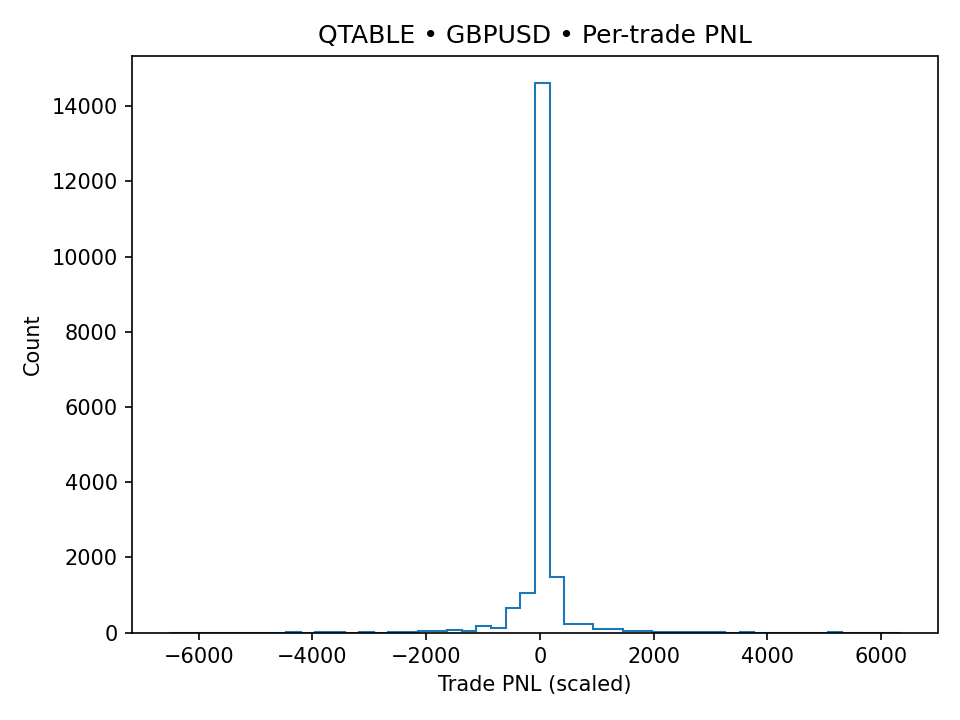}
        \caption{QTable - Histogram}
        \label{fig:aapl_1a}
    \end{subfigure}
    \begin{subfigure}[b]{0.23\textwidth}
        \centering
        \includegraphics[width=\textwidth]{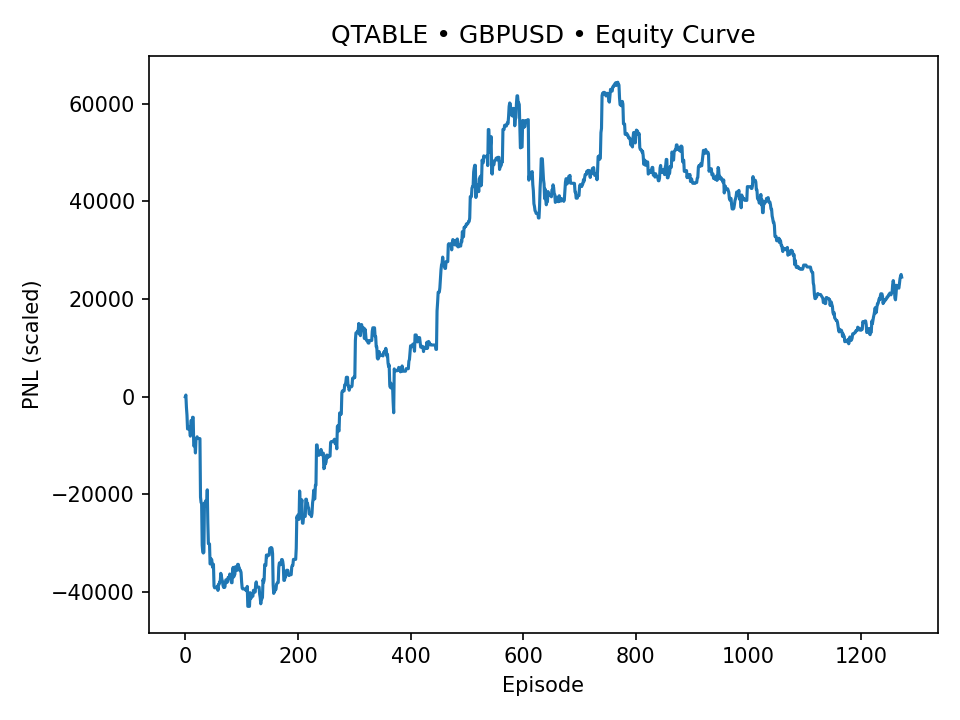}
        \caption{QTable - Equity}
        \label{fig:aapl_1b}
    \end{subfigure}
    \begin{subfigure}[b]{0.23\textwidth}
        \centering
        \includegraphics[width=\textwidth]{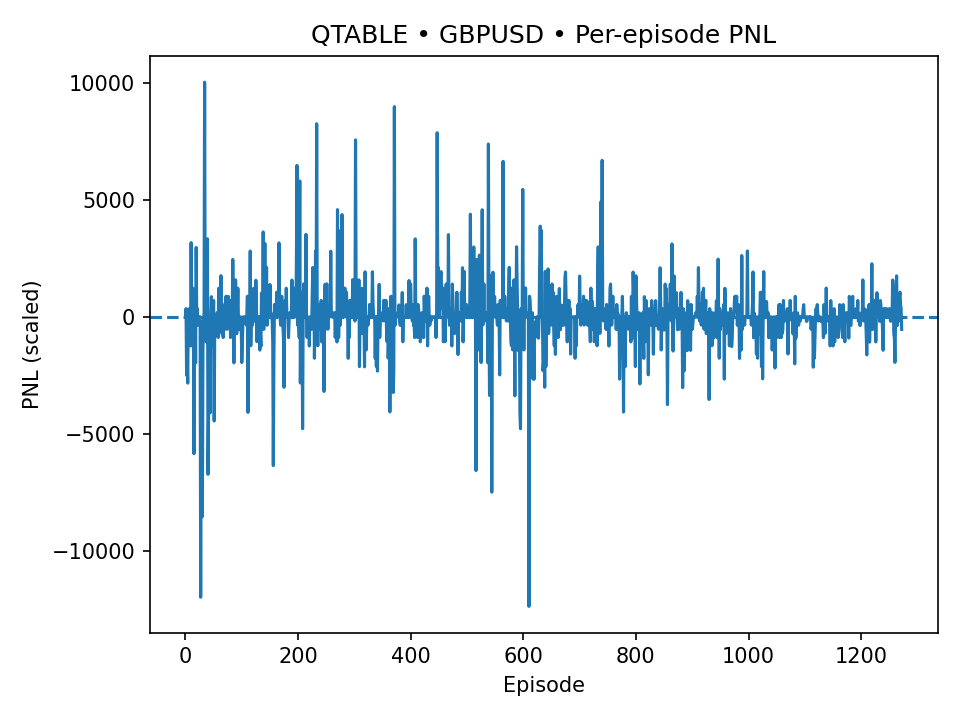}
        \caption{QTable - Returns}
        \label{fig:aapl_1c}
    \end{subfigure}
    \begin{subfigure}[b]{0.23\textwidth}
        \centering
        \includegraphics[width=\textwidth]{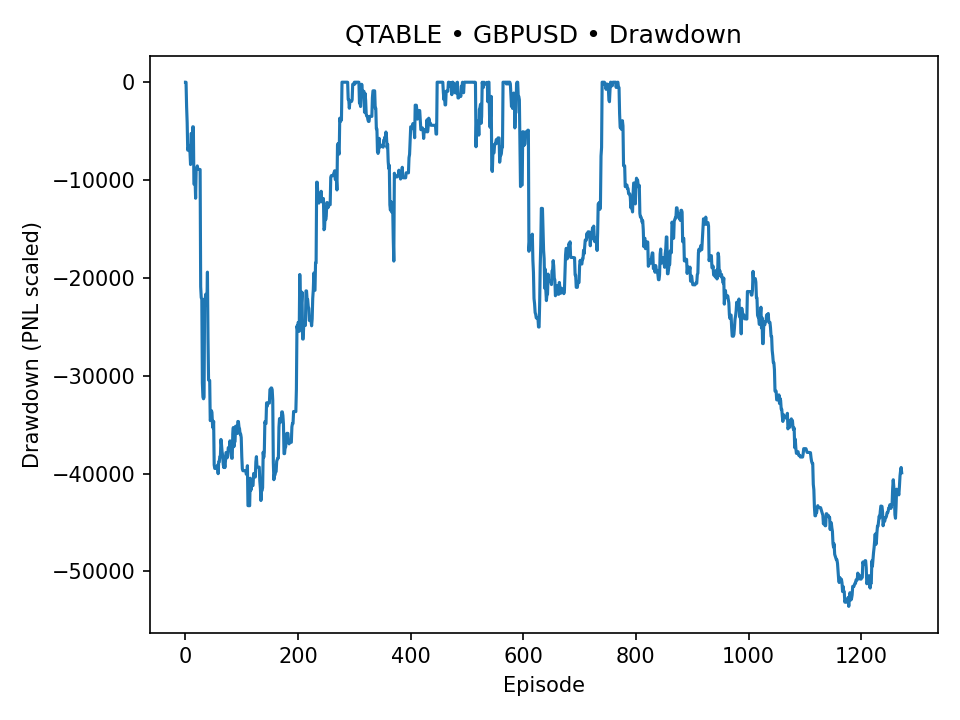}
        \caption{QTable - Drawdown}
        \label{fig:aapl_1d}
    \end{subfigure}
    \\[0.8em]

    \begin{subfigure}[b]{0.23\textwidth}
        \centering
        \includegraphics[width=\textwidth]{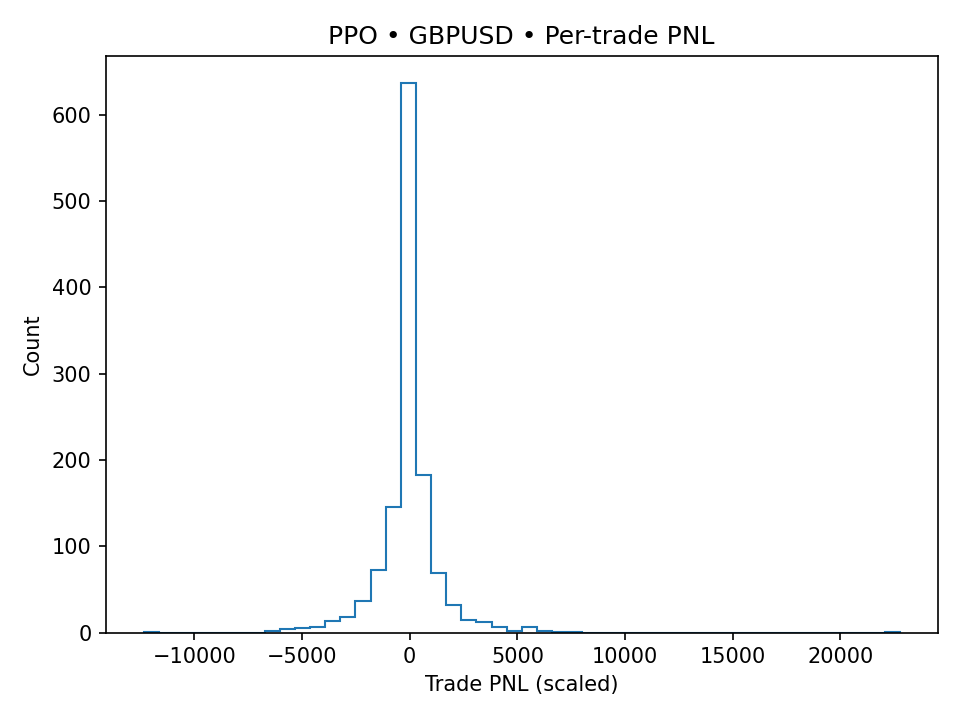}
        \caption{PPO - Histogram}
        \label{fig:aapl_1e}
    \end{subfigure}
    \begin{subfigure}[b]{0.23\textwidth}
        \centering
        \includegraphics[width=\textwidth]{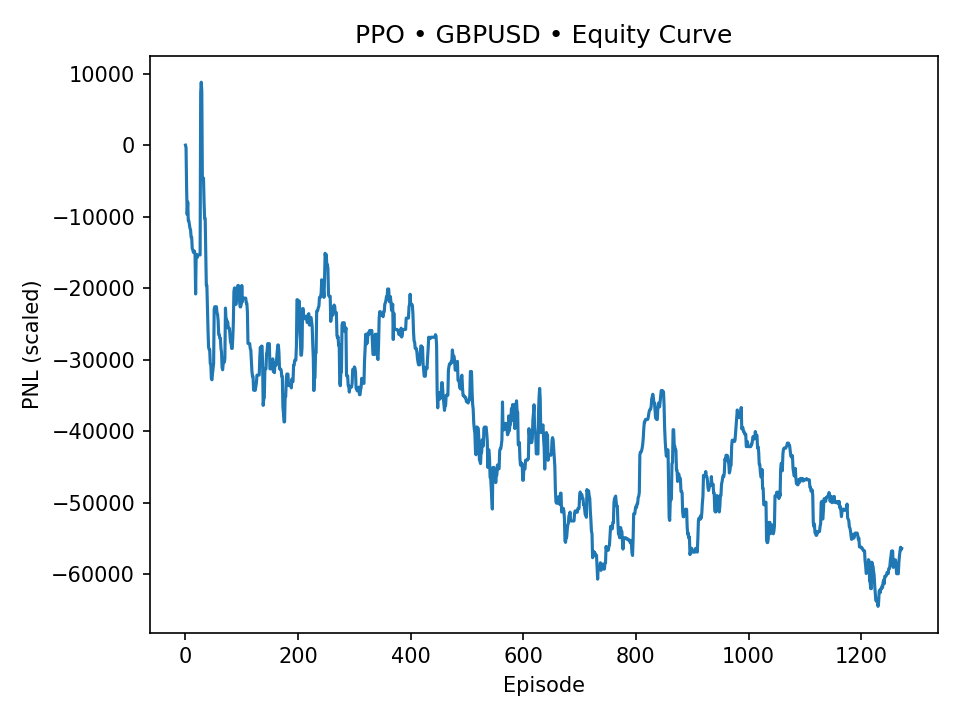}
        \caption{PPO - Equity}
        \label{fig:aapl_1f}
    \end{subfigure}
    \begin{subfigure}[b]{0.23\textwidth}
        \centering
        \includegraphics[width=\textwidth]{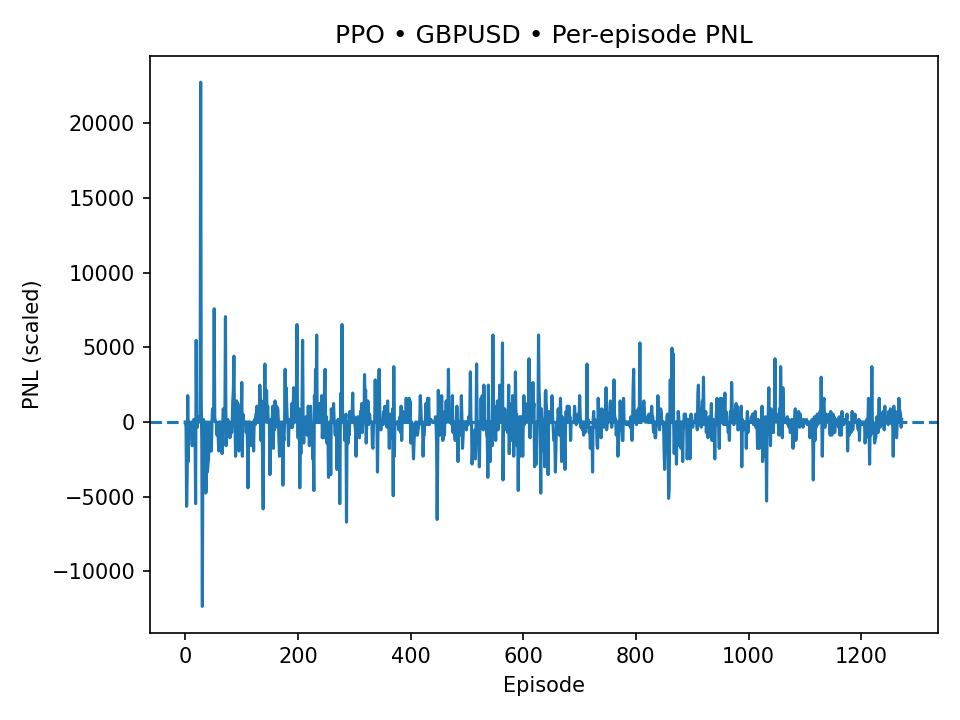}
        \caption{PPO - Returns}
        \label{fig:aapl_1g}
    \end{subfigure}
    \begin{subfigure}[b]{0.23\textwidth}
        \centering
        \includegraphics[width=\textwidth]{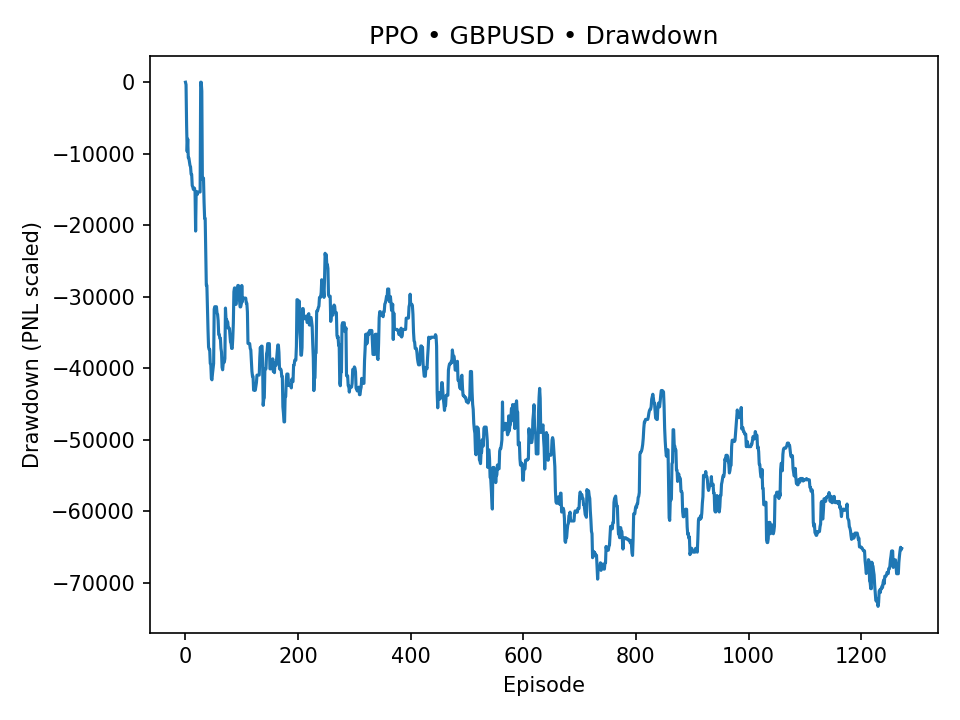}
        \caption{PPO - Drawdown}
        \label{fig:aapl_1h}
    \end{subfigure}
    \\[0.8em]

    \begin{subfigure}[b]{0.23\textwidth}
        \centering
        \includegraphics[width=\textwidth]{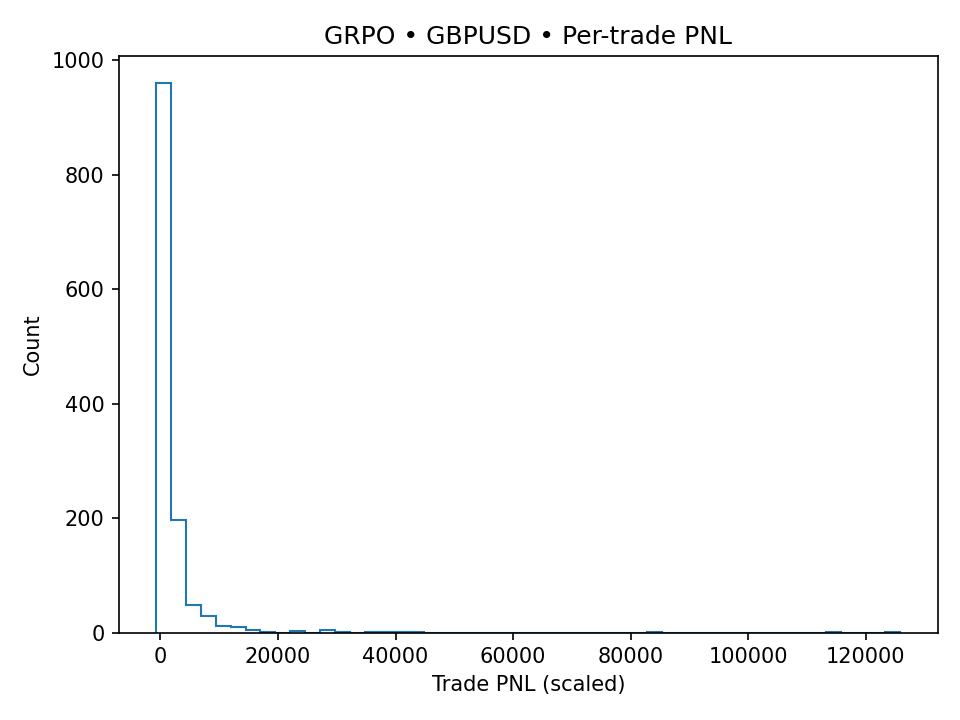}
        \caption{GRPO - Histogram}
        \label{fig:aapl_1i}
    \end{subfigure}
    \begin{subfigure}[b]{0.23\textwidth}
        \centering
        \includegraphics[width=\textwidth]{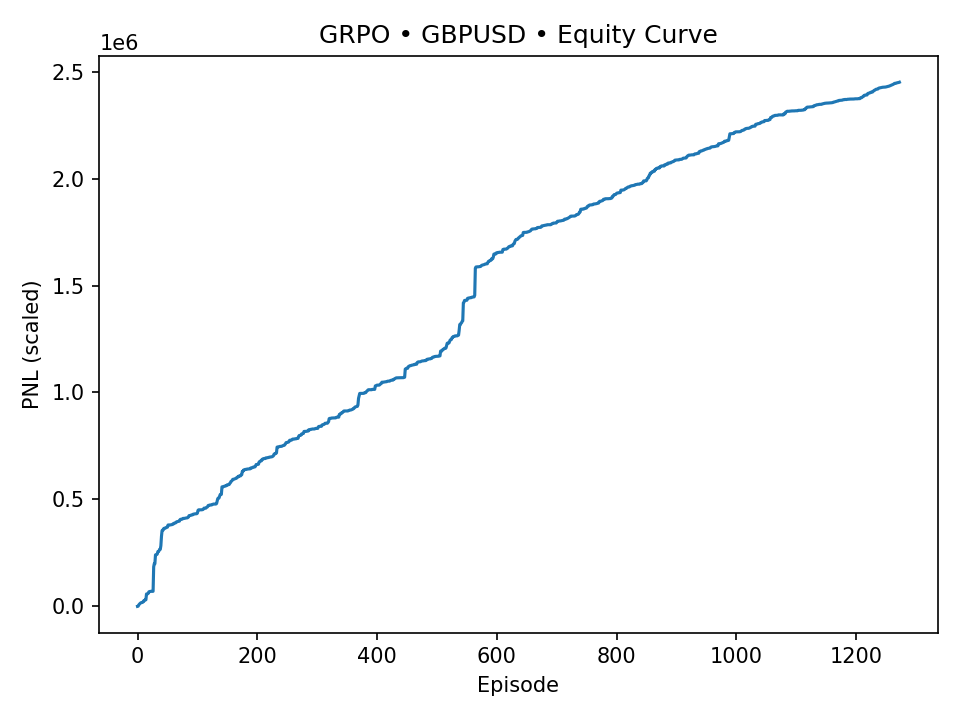}
        \caption{GRPO - Equity}
        \label{fig:aapl_1j}
    \end{subfigure}
    \begin{subfigure}[b]{0.23\textwidth}
        \centering
        \includegraphics[width=\textwidth]{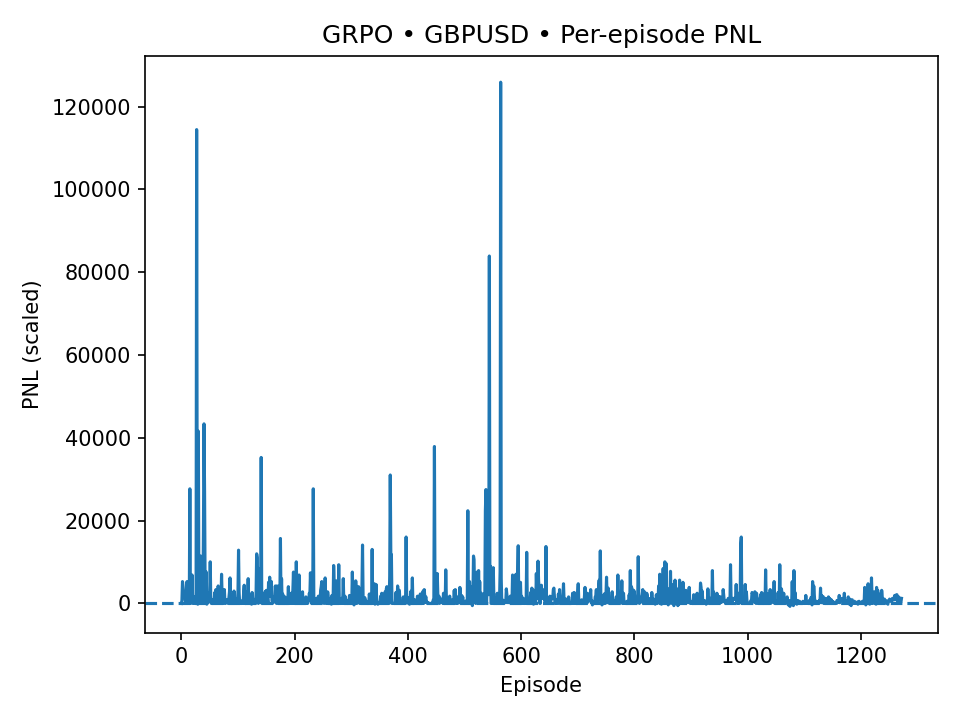}
        \caption{GRPO - Returns}
        \label{fig:aapl_1k}
    \end{subfigure}
    \begin{subfigure}[b]{0.23\textwidth}
        \centering
        \includegraphics[width=\textwidth]{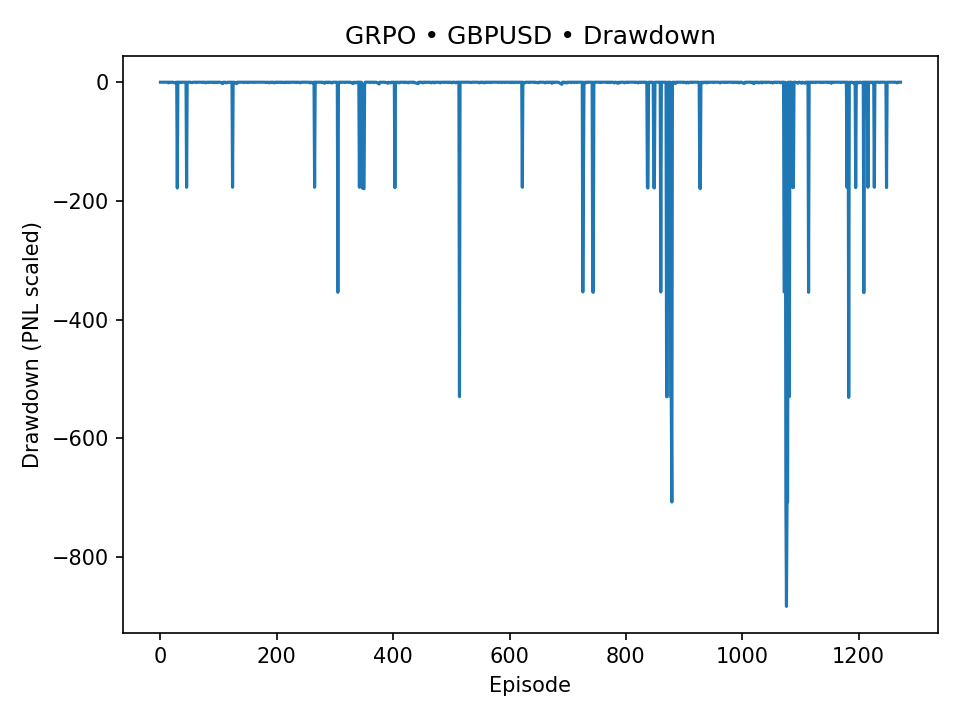}
        \caption{GRPO - Drawdown}
        \label{fig:aapl_1l}
    \end{subfigure}
    \\[0.8em]

    \begin{subfigure}[b]{0.23\textwidth}
        \centering
        \includegraphics[width=\textwidth]{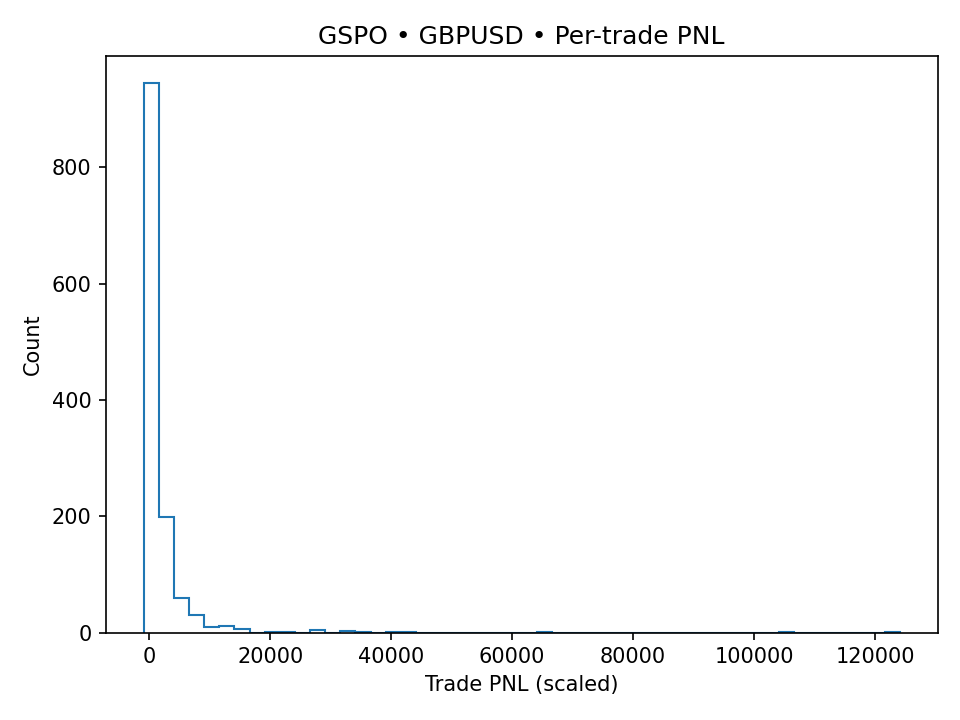}
        \caption{GSPO - Histogram}
        \label{fig:aapl_1m}
    \end{subfigure}
    \begin{subfigure}[b]{0.23\textwidth}
        \centering
        \includegraphics[width=\textwidth]{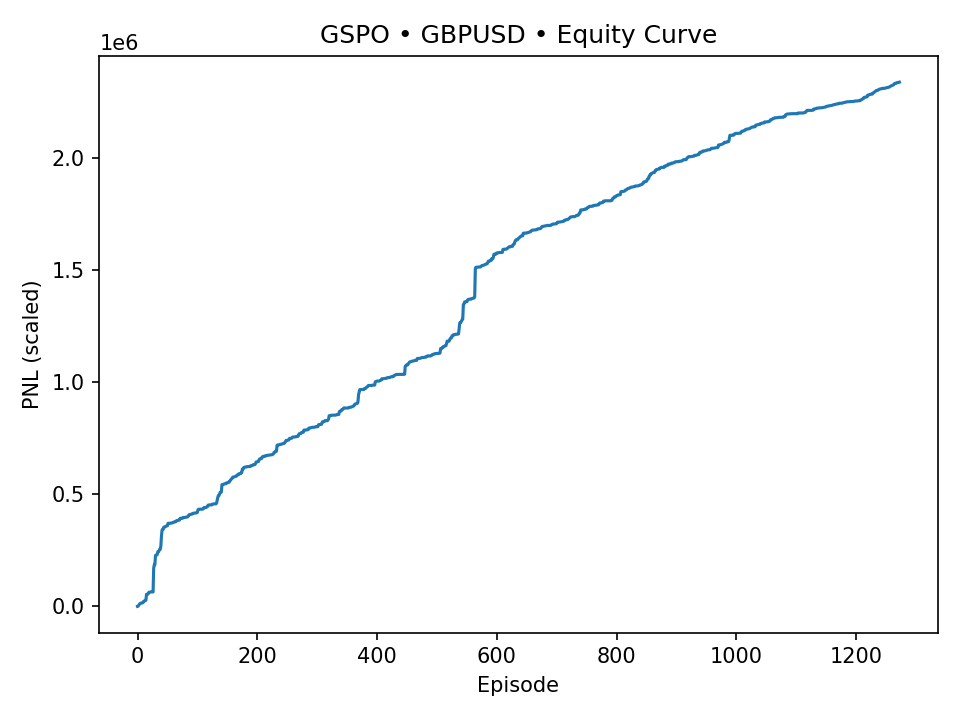}
        \caption{GSPO - Equity}
        \label{fig:aapl_1n}
    \end{subfigure}
    \begin{subfigure}[b]{0.23\textwidth}
        \centering
        \includegraphics[width=\textwidth]{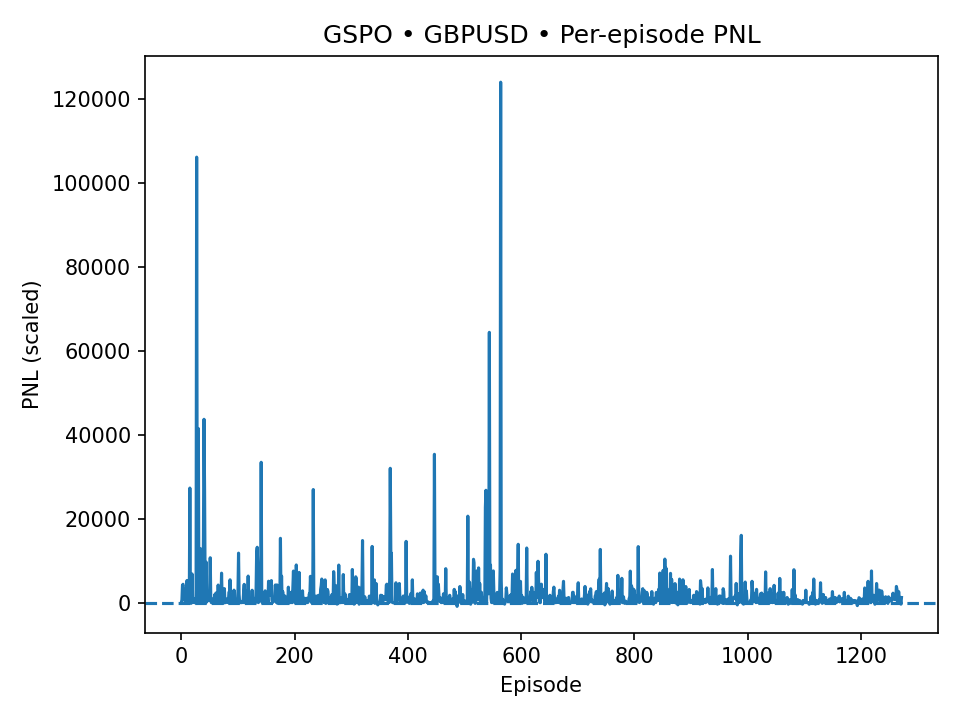}
        \caption{GSPO - Returns}
        \label{fig:aapl_1o}
    \end{subfigure}
    \begin{subfigure}[b]{0.23\textwidth}
        \centering
        \includegraphics[width=\textwidth]{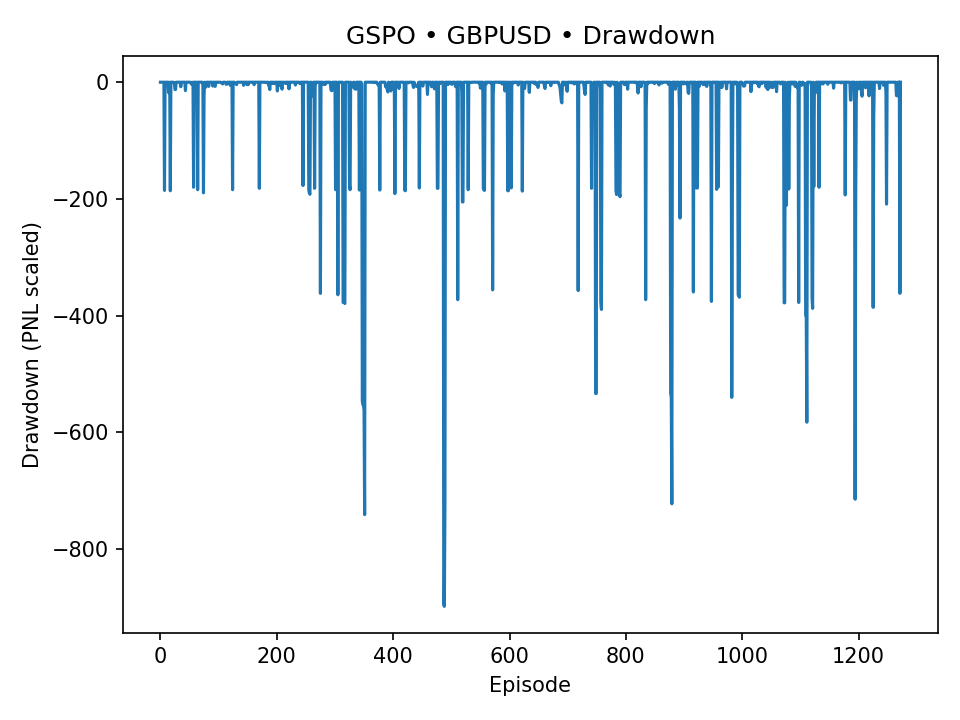}
        \caption{GSPO - Drawdown}
        \label{fig:aapl_1p}
    \end{subfigure}

    \caption{Held-out evaluation plots for AAPL showing the trade-PnL histogram, equity curve (scaled by $10^6$), episode-level PnL returns, and drawdown under the backtesting protocol}
    \label{fig:aapl_4x4}
\end{figure*}

\subsection{Simulation \& Training}

The experimental pipeline has two stages. In the first stage, we train the forecasting network $f_\theta$ for up to 100 epochs with early stopping and retain the best checkpoint according to validation loss. In the second stage, the frozen forecasts $\hat{\alpha}_t$ are used to form the RL state, and the trading policy is trained on the training split of episodes.

\paragraph{Tabular Q-learning.}
As a value-based baseline, we discretize the continuous forecast vector using a map
\[
\mathcal{I}:\mathbb{R}^{H}\rightarrow \{1,\dots,S\},
\]
which assigns each forecast state to one of $S$ discrete bins. Q-learning then applies standard one-step temporal-difference updates with an $\varepsilon$-greedy exploration schedule during training. At evaluation time, actions are selected greedily from the learned $Q$-table.

\paragraph{PPO.}
For PPO, let
\[
\rho_t(\theta)
=
\frac{\pi_\theta(a_t\mid s_t)}{\pi_{\theta_{\mathrm{old}}}(a_t\mid s_t)}
\]
denote the standard per-step importance ratio, and let $\hat{A}_t$ denote the generalized-advantage estimate at step $t$. The clipped PPO objective is
\[
\begin{aligned}
L^{\mathrm{PPO}}(\theta)
= \mathbb{E}_t
\left[
\min\left(
\rho_t(\theta)\hat{A}_t,\;
\operatorname{clip}(\rho_t(\theta),1-\epsilon,1+\epsilon)\hat{A}_t
\right)
\right].
\end{aligned}
\]
We use the standard PPO loss with value-function and entropy regularization.

\paragraph{GRPO.}
To adapt GRPO to trading, each policy update is formed from a batch of $G$ complete episodes
\[
\{\tau_1,\dots,\tau_G\}.
\]
For episode $\tau_i$, define the episode return
\[
R_i = \sum_{t\in \tau_i} r_t,
\qquad
\bar{R} = \frac{1}{G}\sum_{i=1}^{G}R_i.
\]
We then define the update-level group advantage by
\[
A_i^{\mathrm{grp}} = R_i-\bar{R}
\]
or, when standardized in implementation,
\[
A_i^{\mathrm{grp}} = \frac{R_i-\bar{R}}{\sigma_R+\varepsilon},
\qquad
\sigma_R^2 = \frac{1}{G}\sum_{i=1}^{G}(R_i-\bar{R})^2.
\]
Unlike LLM-style GRPO, where grouping is performed over multiple responses to the same prompt, grouping here is performed over the batch of episodes collected in one policy update. The baseline is therefore an update-level empirical baseline rather than a prompt-conditional baseline. The resulting clipped surrogate is applied across the pooled timesteps of the grouped episodes.

\paragraph{GSPO.}
GSPO uses sequence-level rather than per-timestep importance ratios. For each episode $\tau_i$, define
\[
\rho_i(\theta)
=
\exp\left(
\sum_{t\in \tau_i}\log \pi_\theta(a_t\mid s_t)
-
\sum_{t\in \tau_i}\log \pi_{\theta_{\mathrm{old}}}(a_t\mid s_t)
\right)
\]

Using the same grouped episode-level advantage $A_i^{\mathrm{grp}}$, the GSPO objective is
\[
\begin{aligned}
L^{\mathrm{GSPO}}(\theta)
= \frac{1}{G}\sum_{i=1}^{G}
\min\left(
\rho_i(\theta)A_i^{\mathrm{grp}},
\operatorname{clip}(\rho_i(\theta),1-\epsilon,1+\epsilon)A_i^{\mathrm{grp}}
\right)
\end{aligned}
\]
Hence, GSPO constrains the policy update at the level of full trading trajectories rather than individual decision steps.

\paragraph{Interpretation of the group-aware objectives.}
We do not claim a theoretical explanation for why GRPO or GSPO should outperform PPO in trading. However, their adaptation suggests a plausible mechanism. Update-level centering in GRPO reduces sensitivity to the absolute scale of returns within a batch, while sequence-level clipping in GSPO constrains policy updates at the full-trajectory level rather than only locally per action. In noisy episodic trading environments, these design choices may reduce the influence of unusually good or unusually bad episodes on the update. We present this as a mechanistic interpretation of the objectives, not as a theorem or causal proof.

\begin{table*}[!ht]
\centering
\begin{tabular}{llccccc}
\toprule
\textbf{Ticker} & \textbf{Method} & \textbf{Avg Return} & \textbf{Volatility} & \textbf{Avg P/L} & \textbf{Profitability} & \textbf{Max DD} \\
\midrule
\multirow{4}{*}{AMZN}
 & Q-Learning & -10.40 & 2061.59 & 7.99 & 10.90 & -92352.93 \\
 & PPO         & -49.31 & 1567.98 & 2.16 & 29.06 & -9293.30 \\
 & GRPO        & 1796.64 & 10745.69 & \textbf{5341.55} & 62.12 & -715.42 \\
 & GSPO        & \textbf{1815.45} & 10864.15 & 767.52 & 62.24 & -733.42 \\
\midrule
\multirow{4}{*}{AAPL}
 & Q-Learning & 19.24 & 1404.80 & 5.61 & 15.41 & -53564.33 \\
 & PPO         & -44.33 & 1538.85 & 1.66 & 34.98 & -12346.35 \\
 & GRPO        & \textbf{1925.01} & 6408.17 & \textbf{97.69} & 72.86 & -1235.17 \\
 & GSPO        & 1835.91 & 6059.83 & 51.82 & 70.82 & -1428.34 \\
\midrule
\multirow{4}{*}{GOOG}
 & Q-Learning & -63.75 & 2462.16 & 4.16 & 18.46 & -94733.55 \\
 & PPO         & -5.25 & 1498.13 & 2.02 & 32.73 & -8364.61 \\
 & GRPO        & 2069.98 & 5575.11 & \textbf{668.53} & 68.86 & -749.61 \\
 & GSPO        & \textbf{2077.68} & 5721.89 & 187.62 & 66.86 & -1128.92 \\
\bottomrule
\end{tabular}
\caption{Backtest results over the held-out one-hour test window for AMZN, AAPL and GOOG showing the average PnL returns, episode volatility, average Profit-Loss ratio, Profitability \%, and Max Drawdown using different RL methods like Q-Learning, PPO, GRPO and GSPO.}
\label{tab:bt_combined}
\end{table*}

\subsection{Performance Metrics}
Given $E$ episode returns ${\mathcal{R}^{(e)}}*{e=1}^E$ and $M$ trade returns ${g_j}*{j=1}^M$, and instrument scaling $c_{\text{instr}}=\mathrm{PRICE\_TO\_PNL}[\text{instrument}]$, we report
\[
\begin{aligned}
\text{Episode average return} \
&= \frac{1}{E}\sum_{e=1}^E c_{\text{instr}}\mathcal{R}^{(e)}
\end{aligned}
\]
\[
\begin{aligned}
\text{Episode volatility} \
&= \sqrt{\frac{1}{E}\sum_{e=1}^E\bigl(c_{\text{instr}}\mathcal{R}^{(e)}-\overline{R}\bigr)^2}
\end{aligned}
\]
\[
\begin{aligned}
\text{Average profit/loss ratio} \
&= \frac{\mathbb{E}[g_j\mid g_j>0]}{\bigl|\mathbb{E}[g_j\mid g_j<0]\bigr|}
\end{aligned}
\]
\[
\begin{aligned}
\text{Profitability (\%)} \
&= 100\cdot\frac{\#{j:g_j>0}}{\#{j:g_j\neq 0}}
\end{aligned}
\]
Maximum drawdown is the minimum cumulative PnL observed across episodes scaled by $c_{\text{instr}}$. This differs from the conventional peak-to-trough maximum drawdown and should be interpreted as a downside-risk proxy.

\section{Results}
Table~\ref{tab:bt_combined} and Figures~\ref{fig:goog_4x4}, \ref{fig:amzn_4x4}, and \ref{fig:aapl_4x4} summarize the held-out backtest results for AMZN, AAPL, and GOOG under the protocol described in the previous section. Under this setup, GRPO and GSPO achieve higher average return and profitability and lower observed maximum drawdown than PPO and tabular Q-learning across all three tickers. The magnitude of the improvement is substantial in the reported tables: while Q-learning and PPO produce negative or near-zero average returns for several assets and exhibit much larger drawdowns, the group-aware methods produce positive average returns together with markedly smaller worst-case drawdowns. Because we do not report repeated-seed uncertainty, bootstrap confidence intervals, or additional market regimes, the numerical gaps should be viewed as indicative rather than conclusive.

These results should be interpreted as empirical observations under a fixed experimental design rather than as a general statement of method superiority. In particular, the results are reported for the selected ticker-date samples, the chosen train/validation/test partition, the fixed forecasting front-end, and the simplified reward and execution assumptions described earlier. We therefore view the comparison as evidence that the group-aware objectives are promising in this setting, but not as a claim that they are universally superior across assets, regimes, or trading environments.

A tentative interpretation is that update-level centering in GRPO and sequence-level clipping in GSPO may yield more conservative or less variance-sensitive policy updates than vanilla PPO in this noisy episodic setting. However, because the present study does not include repeated-seed uncertainty estimates, training-curve diagnostics, or ablation experiments, this interpretation remains suggestive rather than definitive.

\section{Future Work}

Future work should focus on improving robustness, interpretability, and market realism in a way that makes the proposed framework easier to extend and validate. A first step is to evaluate GRPO and GSPO on a broader universe of assets, over longer sample periods, and across distinct market regimes, while also reporting repeated-seed results, confidence intervals, and sensitivity to train/validation/test splits. This would help determine whether the gains observed here persist under stronger distribution shift and are not tied to a narrow experimental setting.

A second direction is to better understand \emph{why} the group-aware objectives help. In particular, future studies should compare training curves, convergence speed, and reward variance across PPO, GRPO, and GSPO, and include ablations that isolate the effect of group normalization, sequence-level clipping, and reward shaping. Such analysis would make it easier to identify which components are responsible for improved stability and downside control.

Another important extension is to make the simulator more realistic by incorporating explicit transaction costs, slippage, latency, and inventory-related penalties. Studying performance under progressively stronger frictions would provide a clearer picture of how robust these methods remain when the environment more closely resembles practical high-frequency trading.

Finally, future work should evaluate the alpha forecaster as a standalone module and compare the RL agents against simpler forecast-driven baselines, such as threshold or sign-based policies. It would also be valuable to enrich the state with longer-horizon order-book dynamics and inventory information, and to expand the action space and reward design toward richer directional trading objectives rather than directional trading alone.

\bibliographystyle{unsrt}
\bibliography{references}

\end{document}